\documentclass[10pt,journal,compsoc]{IEEEtran}

\usepackage{cite}
\usepackage{multirow}
\usepackage{amssymb}
\usepackage{xspace}
\usepackage{epsf}
\usepackage{setspace}
\usepackage{caption}
\usepackage{comment}
\usepackage{color}
\usepackage{subcaption}
\usepackage[linesnumbered,ruled,vlined]{algorithm2e}
\usepackage{booktabs} 
\usepackage{algpseudocode} 
\usepackage{balance}
\usepackage{url,epsfig,array}
\usepackage{amsmath}
\usepackage{epstopdf}
\usepackage{cases}
\usepackage{bm}
\usepackage{color}
\usepackage{tikz}
\usepackage{threeparttable}
\usepackage{pifont}

\newcommand*{\circled}[1]{\lower.7ex\hbox{\tikz\draw (0pt, 0pt)%
    circle (.5em) node {\makebox[1em][c]{\small #1}};}}

\begin{document}
\title{\huge{Mitigating Catastrophic Forgetting with Adaptive Transformer Block Expansion in Federated Fine-Tuning}}
 \author{
 Yujia Huo,
 Jianchun Liu,~\IEEEmembership{Member,~IEEE,}~
       Hongli Xu,~\IEEEmembership{Member,~IEEE,}~ 
       Zhenguo Ma,~\IEEEmembership{Member,~IEEE,}~
        Shilong Wang,~
          Liusheng Huang,~\IEEEmembership{Member,~IEEE}
 	\IEEEcompsocitemizethanks{
 		\IEEEcompsocthanksitem Y. Huo
   are with the School of Computer Science and Technology, University of Science and Technology of China, Hefei, Anhui, China, 230027, and also with Suzhou Institute for Advanced Research, University of Science and Technology of China, Suzhou, Jiangsu, China, 215123. E-mails: yujia$\_$huo@mail.ustc.edu.cn; jcliu17@ustc.edu.cn; xuhongli@ustc.edu.cn; shilongwang@mail.ustc.edu.cn; lshuang@ustc.edu.cn.
  \IEEEcompsocthanksitem Z. Ma is with the School of Computer Science and Technology, China University of Mining and Technology, Xuzhou, Jiangsu, China, 221116, and also with Mine Digitization Engineering Research Center of the Ministry of Education. E-mail: cs$\_$zgma@cumt.edu.cn.
 	}
 }

\markboth{IEEE Transactions on Mobile Computing}%
{Shell \MakeLowercase{\textit{et al.}}: Bare Advanced Demo of IEEEtran.cls for Journals}

\IEEEtitleabstractindextext{
\begin{abstract}
Federated fine-tuning (FedFT) of large language models (LLMs) has emerged as a promising solution for adapting models to distributed data environments while ensuring data privacy.
 Existing FedFT methods predominantly utilize parameter-efficient fine-tuning (PEFT) techniques to reduce communication and computation overhead. 
 However, they often fail to adequately address the catastrophic forgetting, a critical challenge arising from continual adaptation in distributed environments.
The traditional centralized fine-tuning methods, which are not designed for the heterogeneous and privacy-constrained nature of federated environments, struggle to mitigate this issue effectively.
Moreover, the challenge is further exacerbated by significant variation in data distributions and device capabilities across clients, which leads to intensified forgetting and degraded model generalization.
To tackle these issues, we propose FedBE, a novel FedFT framework that integrates an adaptive transformer block expansion mechanism with a dynamic trainable-block allocation strategy. 
Specifically, FedBE expands trainable blocks within the model architecture, structurally separating newly learned task-specific knowledge from the original pre-trained representations.
Additionally, FedBE dynamically assigns these trainable blocks to clients based on their data distributions and computational capabilities. This enables the framework to better accommodate heterogeneous federated environments and enhances the generalization ability of the model.
 Extensive experiments show that compared with existing federated fine-tuning methods, FedBE achieves 12–74\% higher accuracy retention on general tasks after fine-tuning and a model convergence acceleration ratio of 1.9-3.1$\times$ without degrading the accuracy of downstream tasks.
\end{abstract}
}

\maketitle
\IEEEdisplaynontitleabstractindextext
\IEEEpeerreviewmaketitle

\section{Introduction}\label{sec:intro}
Recently, transformer-based large language models (LLMs) have achieved remarkable success in NLP through the pre-training and fine-tuning paradigm \cite{devlin2019bert, brown2020language}, but this approach typically requires centralized access to large-scale labeled data, raising privacy concerns under regulations like GDPR \cite{shokri2015privacy}. To address this, federated fine-tuning (FedFT) has emerged as a privacy-preserving alternative, enabling decentralized clients to adapt models without sharing raw data. FedNLP \cite{lin2021fednlp} exemplifies this approach by collaboratively optimizing full model parameters. However, full-parameter tuning incurs heavy communication and computation overhead, making it unsuitable for resource-constrained clients.
To alleviate these burdens, researchers have introduced parameter-efficient fine-tuning (PEFT) techniques \cite{houlsby2019parameter} into the FedFT setting. For instance, FedAdapter \cite{cai2022fedadapter} and FedPETuning \cite{zhang2023fedpetuning} incorporate lightweight modules such as adapters, LoRA, and BitFit to reduce the volume of updated parameters during training. By freezing the pre-trained LLM backbone and only updating inserted parameters, PEFT methods significantly reduce computation overhead on the client side. Moreover, as only these modules are transmitted between clients and the server, the communication cost is also greatly reduced compared to full-model updates.

Although recent advances in FedFT have yielded impressive improvements in task performance and training efficiency, \textit{catastrophic forgetting} \cite{de2021continual} remains an underexplored problem. 
This phenomenon manifests as the model is adapted to new tasks and forgets previously acquired knowledge, resulting in significant performance degradation on previous tasks.
While catastrophic forgetting has been extensively studied in centralized continual learning, with techniques such as memory replay \cite{hayes2020remind} and regularization-based methods (e.g., R-EWC \cite{liu2018rotate}), these solutions do not directly translate to the federated settings. 
Memory replay mitigates forgetting by storing and revisiting samples from prior tasks, but such data retention is incompatible with privacy constraints in federated environments. 
Besides, methods like R-EWC utilize Fisher information matrices computed from gradients of previous tasks to penalize deviations in critical parameters. However, in federated fine-tuning, each client typically has access only to a fragmented subset of data, resulting in inaccurate Fisher matrix estimation and weakened forgetting mitigation \cite{karakida2019universal}.
In the context of federated learning, several studies have explored mitigating catastrophic forgetting by introducing regularization mechanisms or decoupled training strategies. These approaches typically aim to preserve global knowledge while allowing local adaptation.
For instance, FedCurv \cite{shoham2019overcoming} adds a Fisher-based regularization term to constrain local model updates and reduce forgetting. However, when applied to the fine-tuning of large language models, such constraints may overly limit parameter updates, leading to underfitting and poor downstream generalization.

Furthermore, we observe that the catastrophic forgetting issue is exacerbated in federated fine-tuning, primarily due to the inherent challenge of \textit{data heterogeneity},
which arise from differences in user behaviors, environmental contexts, or application-specific requirements~\cite{zhao2018federated,kairouz2019advances}.
As a result, locally fine-tuned model parameters tend to diverge significantly from one another, impairing their ability to capture and generalize global knowledge \cite{marfoq2021federated}. 
Besides, such parameter divergence hinders the effectiveness of model aggregation, making it difficult to preserve the original model's generalization capacity and exacerbating catastrophic forgetting.
As demonstrated in Section \ref{subsec:limitation}, under identical training hyperparameters, the accuracy of general classification tasks decreases by 11–14\% and 17-27\% in centralized and federated settings respectively, after fine-tuning on specific downstream tasks. 
Moreover, as the heterogeneity of data distribution increases, the model tends to forget more of its general knowledge, which is reflected in a more substantial drop in accuracy on general tasks.
To address the above issues, we propose an efficient FedFT framework named FedBE, which incorporates an adaptive transformer block expansion mechanism. 
Specifically, FedBE expands the base model by appending additional transformer blocks as trainable modules while keeping the original parameters frozen. 
This strategy structurally decouples the model’s general knowledge from downstream task adaptation, thereby isolating new knowledge from existing representations to mitigate catastrophic forgetting.
Moreover, FedBE adapts to data and resource heterogeneity across clients by employing a dynamic trainable block assignment strategy. 
This strategy assigns different subsets of the expanded transformer blocks to each client based on its specific data characteristics and computational resource capacity, enabling flexible adaptation without compromising overall system efficiency.

\begin{figure}[t] \centering
    \includegraphics[width=0.5\textwidth]
    {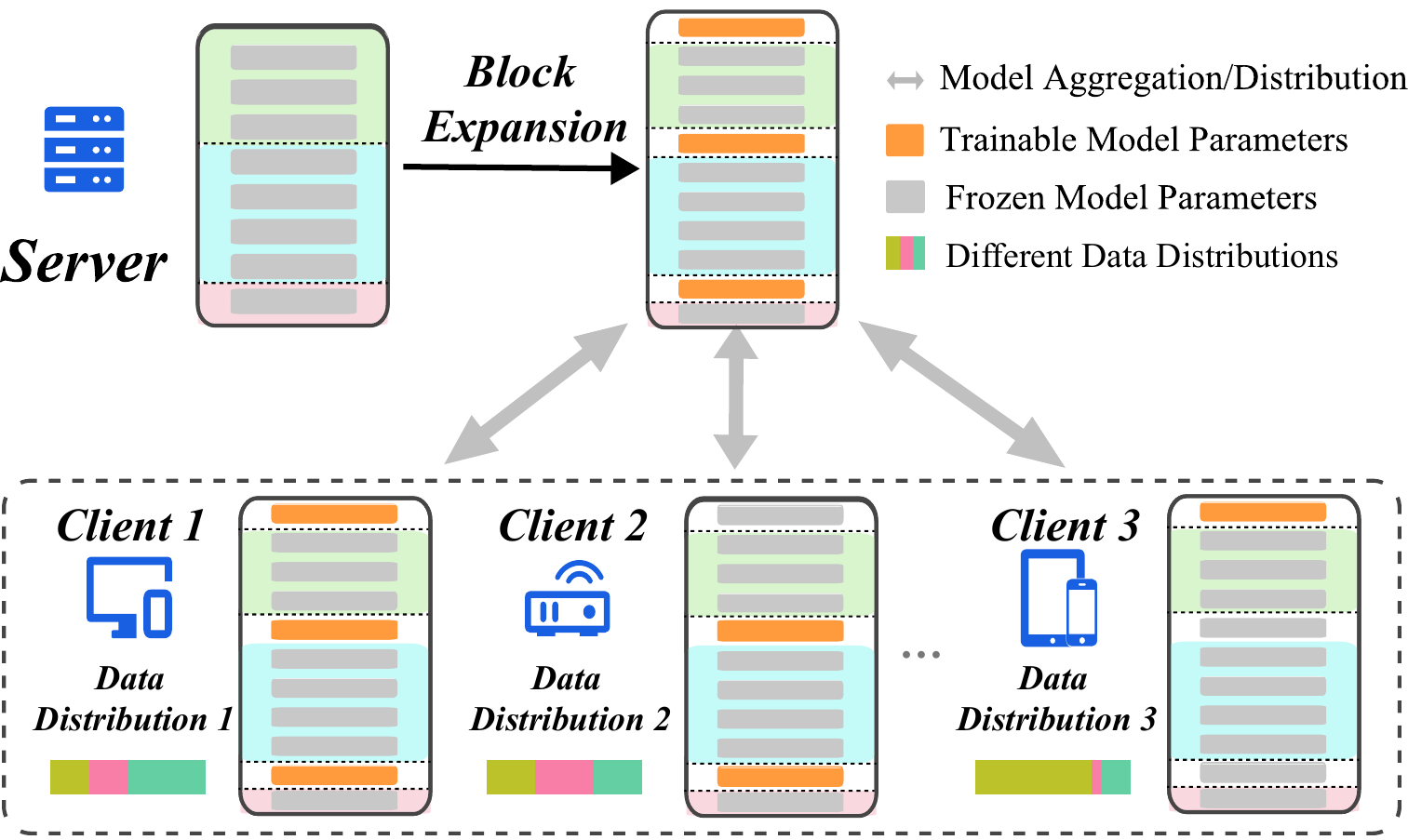}
    \caption{The illustration of transformer block expansion and trainable block assignment in FedBE.}\label{fig:overview}
\end{figure}

 In FedBE, as illustrated in Fig.~\ref{fig:overview}, the server expands the base model by appending trainable blocks (orange) to the frozen backbone (gray), and distributes different subsets of these blocks to clients based on their computational resources and data distribution characteristics. For instance, Client 1, equipped with abundant computational resources, is assigned all expanded blocks, allowing it to fully participate in model training. Client 2, with moderate resources and a relatively uniform data distribution, is assigned two lower-layer blocks, enabling it to focus on optimizing the model’s understanding of input-level semantics. In contrast, Client 3, which has the most limited resources and highly skewed data distribution, is assigned only one top-layer block to concentrate on abstract feature analysis.

Compared with PEFT methods such as inserting adapter modules, FedBE introduces a trainable transformer block after each frozen transformer layer.
These blocks are zero-initialized to avoid interfering with the original model and are designed to process inputs in parallel with the frozen transformer blocks, rather than being inserted into it.
This design provides sufficient representational capacity for new tasks, constrains gradient flow within the expanded blocks, and prevents reshaping of the original representations. 
Therefore, FedBE enables a clear structural separation between old and new knowledge, thereby significantly mitigating catastrophic forgetting-a challenge that adapter-based methods, constrained by their tight integration with the backbone and limited representational capacity, often fail to address effectively.
In summary, FedBE not only enhances the model’s ability to retain general knowledge during local fine-tuning, but also ensures efficient and personalized adaptation across heterogeneous clients.

Despite its effectiveness, the performance of FedBE critically depends on two key design factors: (1) determining the proper number and position of expanded blocks, and (2) designing an effective trainable block allocation strategy for heterogeneous clients.
Firstly, while increasing the number of expanded blocks can enhance downstream task performance, it also introduces additional training and inference overhead. 
Moreover, the position of these blocks within the transformer architecture plays a pivotal role, as different insertion positions can lead to significantly different performance outcomes.
Hence, \textit{it is crucial to jointly optimize both the number and position of expanded blocks} to strike a balance between performance and efficiency.
Secondly, improper trainable blocks allocation across heterogeneous clients may lead to suboptimal fine-tuning performance and exacerbate catastrophic forgetting, due to disparities in data distribution and computational resources. 
This necessitates \textit{the development of a resource-aware and data-sensitive allocation strategy}, which can adaptively distribute training responsibilities across clients, to maximize overall training performance while preserving general knowledge.

Our main contributions are summarized as follows:
\begin{itemize}
    \item We propose a novel federated fine-tuning framework named FedBE, which leverages transformer block expansion and training block assignment, to significantly mitigate catastrophic forgetting and improve the generalization performance of downstream tasks.
    \item We develop an expanded block selection algorithm that determines the number and positions of expanded blocks by analyzing the gradient magnitudes and relative positions of transformer blocks.
    \item We design a greedy-based algorithm that dynamically allocates training blocks to clients according to their data distribution and resource capacity.
    \item We validate our method and baselines on real hardware, demonstrating 12–74\% higher accuracy retention on general tasks after fine-tuning compared to baselines, indicating stronger resistance to catastrophic forgetting and better downstream performance.
\end{itemize}

\section{Background and Motivation}\label{sec:motivation}
\subsection{Federated Fine-Tuning (FedFT)}

The traditional fine-tuning paradigms typically trains the model by centralizing the raw data of edge clients.\cite{devlin2019bert,raffel2020exploring}.
Considering privacy regulations such as GDPR, the centralization of raw data incurs serious risks in real-world applications, especially when user-sensitive information is involved in edge scenarios~\cite{huang2024keystrokesniffer}.
To circumvent this issue, federated fine-tuning (FedFT) \cite{lin2021fednlp} emerges as a promising alternative, allowing multiple decentralized clients to collaboratively fine-tune a shared pre-trained model exchanging the raw data.

FedFT is fundamentally an extension of federated learning \cite{mcmahan2017communication}, focusing specifically on updating a pre-trained global model rather than training it from scratch. Formally, given a federated setup comprising $M$ clients, each client $C_m$, where $m \in \{1,2,\dots,M\}$, holds a local dataset $D_m$ consisting of $n_m$ samples. In the $t$-th round, the server distributes the global model $\theta^{t}$ (where $\theta^{0}$ is the pre-trained model) to clients. Each client locally fine-tunes the model and generates an update $\Delta \theta_{m}^{t}$. Subsequently, the server collects these updates and aggregates them to generate a new global model:

\begin{equation}
\theta^{t+1} = \sum_{m=1}^{M} \frac{n_m}{|S|} \cdot (\theta^{t} + \Delta \theta_{m}^{t}),
\end{equation}
where $|S| = \sum_{m=1}^{M} n_m$ denotes the total number of samples across all clients.

Considering the massive parameter scale of large language models (LLMs), parameter-efficient fine-tuning (PEFT) techniques have been widely adopted in federated fine-tuning (FedFT) to reduce communication overhead and improve training efficiency. As a result, the updates $\Delta \theta_{m}^{t}$ no longer represent full-scale model updates, but rather modifications confined to newly introduced lightweight modules, such as LoRA~\cite{hu2022lora} and Adapter layers~\cite{houlsby2019parameter}. This strategy has led to substantial reductions in latency and network traffic, thereby enhancing the practicality of FedFT in real-world deployments.



\subsection{Limitations of Existing FedFT Methods}\label{subsec:limitation}

Despite these advancements, FedFT still faces critical unresolved challenges, notably catastrophic forgetting exacerbated by data heterogeneity.
As discussed in Section \ref{sec:intro}, catastrophic forgetting is notably intensified in federated fine-tuning compared to traditional centralized training \cite{yang2024federated}.
This intensified forgetting primarily arises due to data heterogeneity and highly variable nature of local training, causing significant divergence between client models and exacerbating performance degradation on previously learned general tasks.
There are some methods for catastrophic forgetting in centralized fine-tuning. For example, traditional memory replay methods \cite{hayes2020remind}, widely successful in centralized continual learning, become infeasible due to strict privacy constraints preventing data sharing across clients. Even if pseudo-sample replay is employed, the inherent data heterogeneity means that replayed data may not accurately represent the global task distribution, thus failing to mitigate forgetting effectively. Similarly, regularization-based approaches such as R-EWC \cite{hayes2020remind} rely heavily on accurate estimations of parameter importance derived from global gradients. However, under federated fine-tuning, each client has access only to fragmented subsets of data, which severely hampers the accuracy of such estimations. Consequently, parameter regularization may excessively constrain locally important task adaptations, impairing client-specific task performance and exacerbating forgetting upon aggregation.

\begin{figure}[t]
    \centering
    \begin{subfigure}[b]{0.49\columnwidth}
        \centering
        \includegraphics[width=\textwidth]{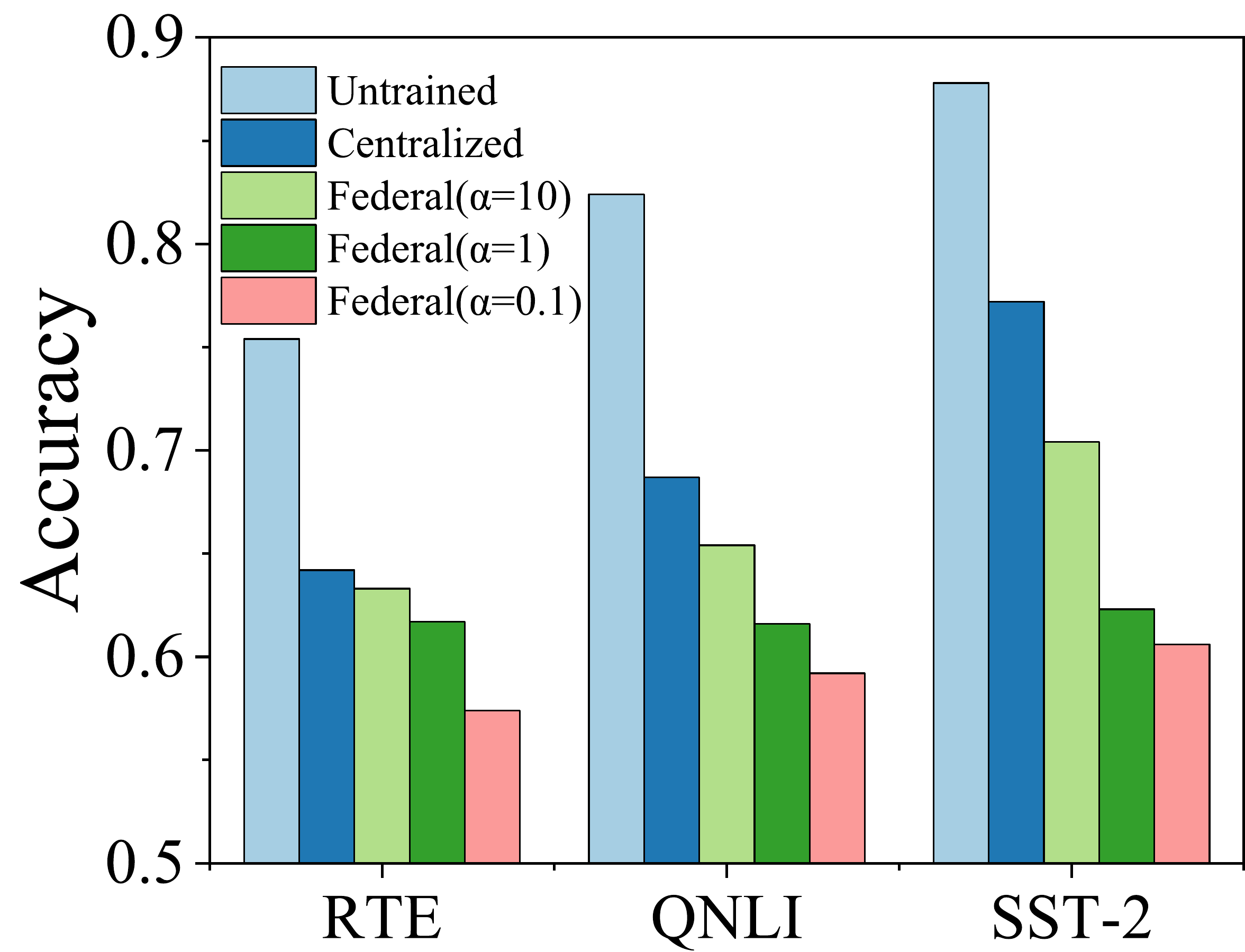}
        \caption{Accuracy under different settings}
        \label{fig:different-environment}
    \end{subfigure}
    \hfill
    \begin{subfigure}[b]{0.49\columnwidth}
        \centering
        \includegraphics[width=\textwidth]{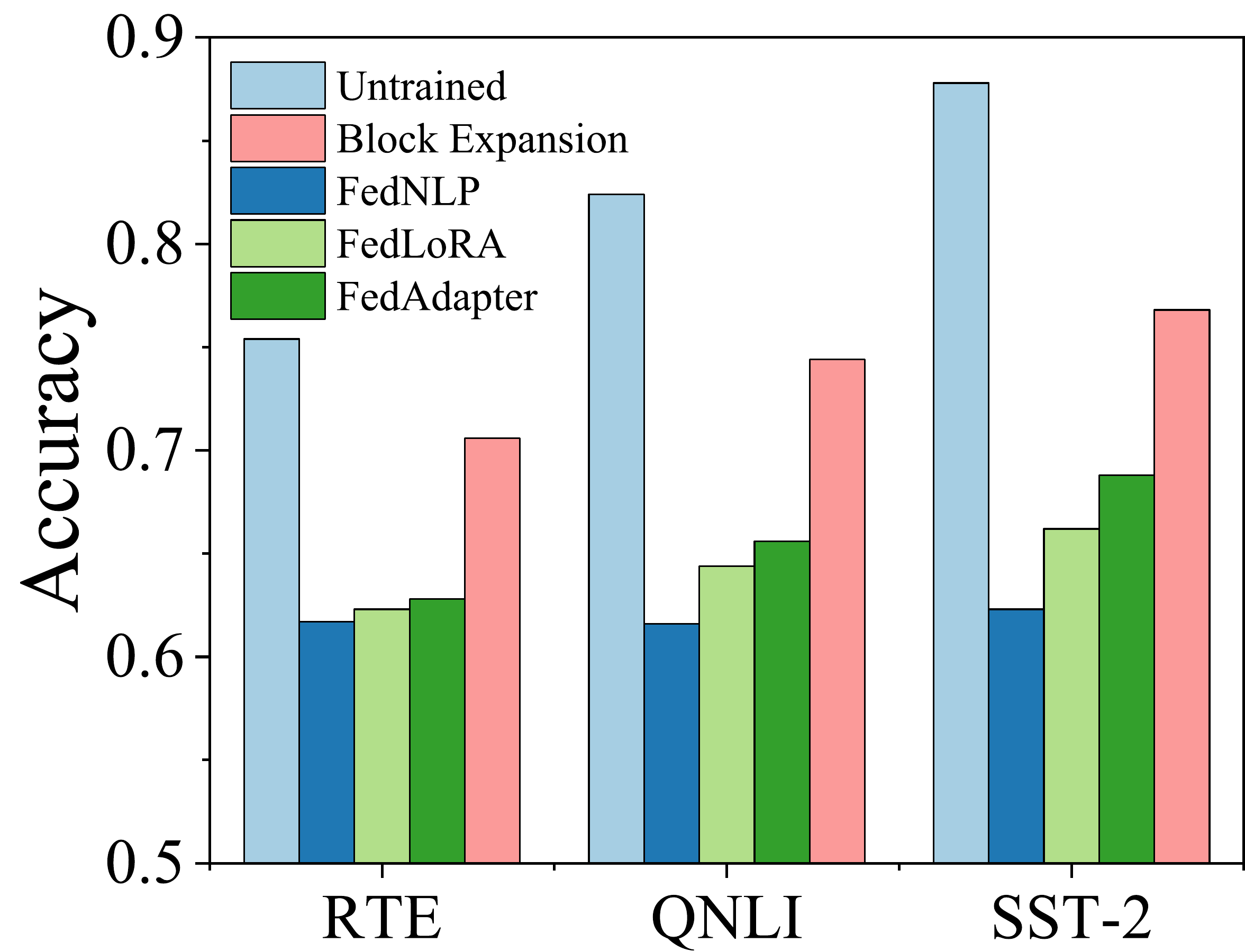}
        \caption{Accuracy under different methods}
        \label{fig:different-method}
    \end{subfigure}
    \caption{Impact of different settings and methods on catastrophic forgetting across RTE, QNLI, and SST-2.}
    \label{fig:catastrophic-comparison}
\end{figure}

In order to systematically demonstrate the catastrophic forgetting problem in FedFT, we conduct experiments based on the RoBerta-base model \cite{liu2019roberta}. Using MRPC (Microsoft Research Paraphrase Corpus) \cite{misra2002mrpc} as the primary task, we compare models fine-tuned in centralized and federated environments by transferring these models and evaluating their performance on general language understanding tasks including RTE (Recognizing Textual Entailment) \cite{dagan2005pascal}, SST-2 (Stanford Sentiment Treebank) \cite{socher2013recursive}, and QNLI (Question Natural Language Inference) \cite{rajpurkar2016squad}. To quantify the degree of catastrophic forgetting, we measure the accuracy variations on these general tasks before and after MRPC fine-tuning.
Specifically, we use the performance of models individually fine-tuned on each general task (i.e., without prior MRPC fine-tuning) as the upper-bound reference, which is denoted as "Untrained" in Fig.~\ref{fig:catastrophic-comparison}. After performing MRPC fine-tuning (centralized or federated), the model is re-evaluated on these tasks. The degradation in accuracy relative to the upper-bound value is used as a quantitative indicator of knowledge forgetting.
In both settings, we adopt a full fine-tuning approach, updating all model parameters during training. In FedFT, we set the number of clients to 10 and simulate varying degrees of non-IID by partitioning data with different configurations of Dirichlet distribution parameter ($\alpha$ = 10.0, 1.0, and 0.1) \cite{hsu2019measuring}, \cite{liu2023finch}. Formally, for a classification task with $K$ classes, the label distribution of client $m$ is sampled from a Dirichlet distribution as follows:
\begin{equation}
\mathbf{p}_m \sim \text{Dir}(\alpha \cdot \mathbf{1}_K),
\end{equation}
where $\mathbf{p}_m = (p_{m,1}, p_{m,2}, \dots, p_{m,K})$ denotes the class distribution vector of client $m$, $\mathbf{1}_K$ is a $K$-dimensional all-one vector, and $\alpha > 0$ is the concentration parameter that controls the degree of non-IIDness. A larger $\alpha$ results in more uniform class distributions across clients (i.e., closer to IID), while a smaller $\alpha$ yields more skewed distributions (i.e., stronger non-IIDness), as clients tend to concentrate on fewer classes.

As depicted in Fig.~\ref{fig:catastrophic-comparison}(a), FedFT consistently results in more severe accuracy degradation compared to centralized fine-tuning across all evaluated general tasks. Moreover, as the data distribution becomes increasingly uneven (i.e., as $\alpha$ decreases), the extent of accuracy degradation also increases, highlighting the sensitivity of FedFT to non-IID data.
Specifically, under the most heterogeneous setting ($\alpha=0.1$), models suffer from accuracy drops ranging from 13\% to 31\% across different general tasks—significantly higher than in the centralized fine-tuning scenario (approximately 4.5\%–12\%). These results clearly demonstrate that FedFT substantially exacerbates catastrophic forgetting, primarily due to severe data heterogeneity and inconsistent global updates.

We further evaluate existing FedFT methods, including FedNLP, FedLoRA \cite{zhang2023fedpetuning}, FedAdapter \cite{cai2022fedadapter}, regarding their effectiveness in mitigating catastrophic forgetting. In these experiments, we use MRPC as the training set to fine-tune the RoBerta-base model under the aforementioned federated setting using different methods, and subsequently evaluate their transfer performance on RTE, SST-2, and QNLI. The results summarized in Fig.~\ref{fig:catastrophic-comparison}(b) indicate that all tested methods exhibit significant accuracy degradation on general tasks after federated fine-tuning. Specifically, FedNLP and FedLoRA, which separately perform full-model and partial low-rank updates, show substantial accuracy drops, averaging around 10\%–29\%. FedAdapter, despite using lightweight modules, still demonstrates noticeable performance deterioration of around 9\%–23\%. 

Our in-depth analysis reveals several critical limitations of these methods, which account for their limited effectiveness in mitigating catastrophic forgetting in federated fine-tuning. Firstly, mainstream methods such as FedNLP typically update all the model parameters, overwriting the general knowledge representations with task-specific gradients. This lack of separation between  general knowledge and newly acquired task-specific knowledge significantly worsens forgetting. Secondly, although parameter-efficient methods like FedAdapter and FedLoRA reduce the volume of parameter updates, they lack mechanisms to account for the varying importance of different layers in specific downstream tasks. Treating all layers equally in these methods limits the model’s ability to adapt to task-specific features, which in turn affects knowledge retention after fine-tuning. Thirdly, most current approaches lack adaptation to address data heterogeneity across clients. Employing a uniform training strategy fails to accommodate diverse local data distributions, resulting in local updates that diverge from global objectives and impairing overall performance after aggregation.


In conclusion, current FedFT methods remain insufficient in mitigating catastrophic forgetting. There is a pressing need for frameworks that incorporate knowledge isolation, task adaptability, and resource-aware strategies to enhance model generalization in federated settings.

\subsection{Opportunity to Mitigate Catastrophic Forgetting}

\begin{figure}[t]
    \centering
    \subcaptionbox{ MRPC accuracy with different numbers of expanded blocks\label{fig:different-environment}}[0.49\columnwidth][c]{%
        \includegraphics[width=\linewidth]{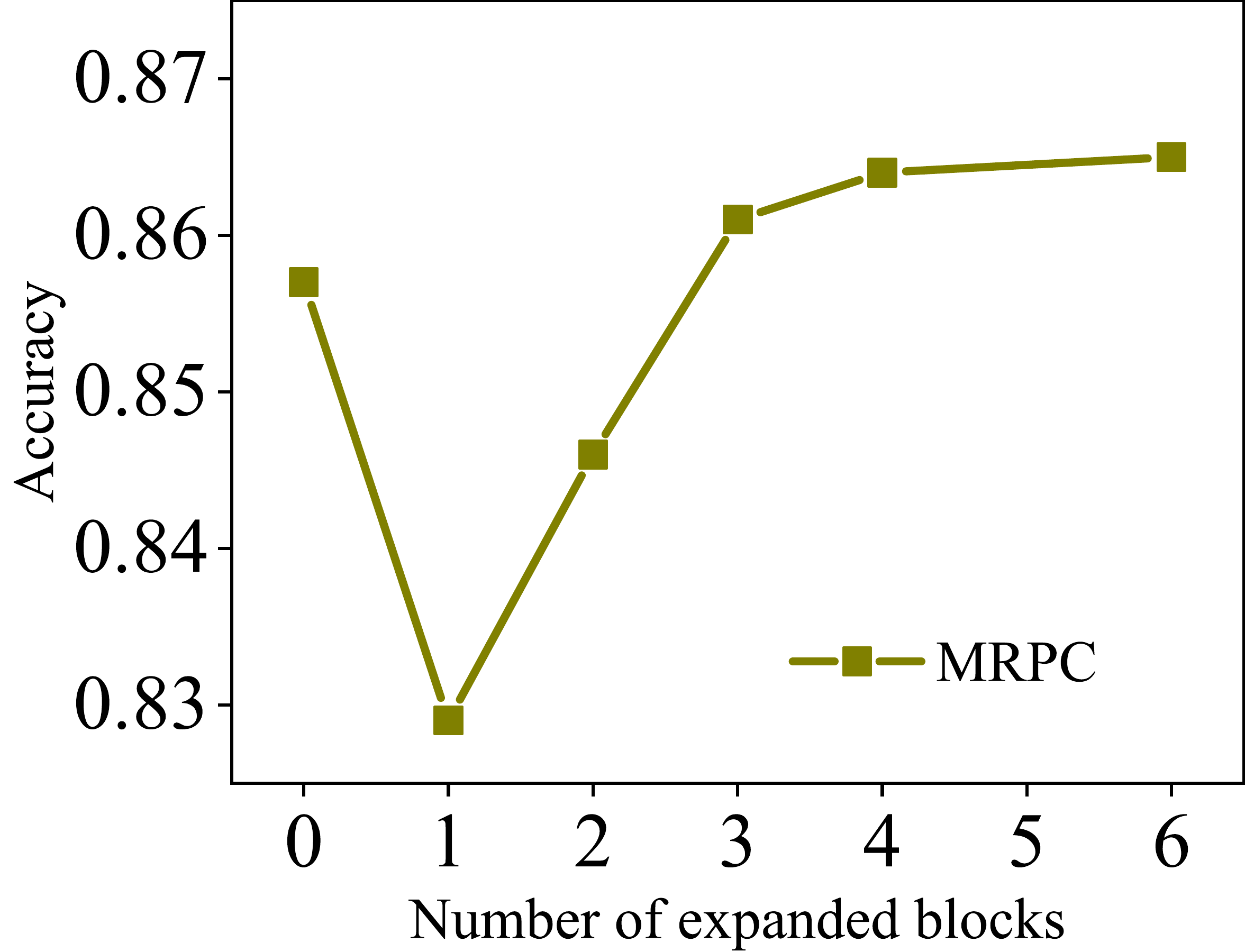}
    }
    \hfill
    \subcaptionbox{ Accuracy decrease on RTE, QNLI, and SST-2 with different numbers of expanded blocks.
\label{fig:different-method}}[0.49\columnwidth][c]{%
        \includegraphics[width=\linewidth]{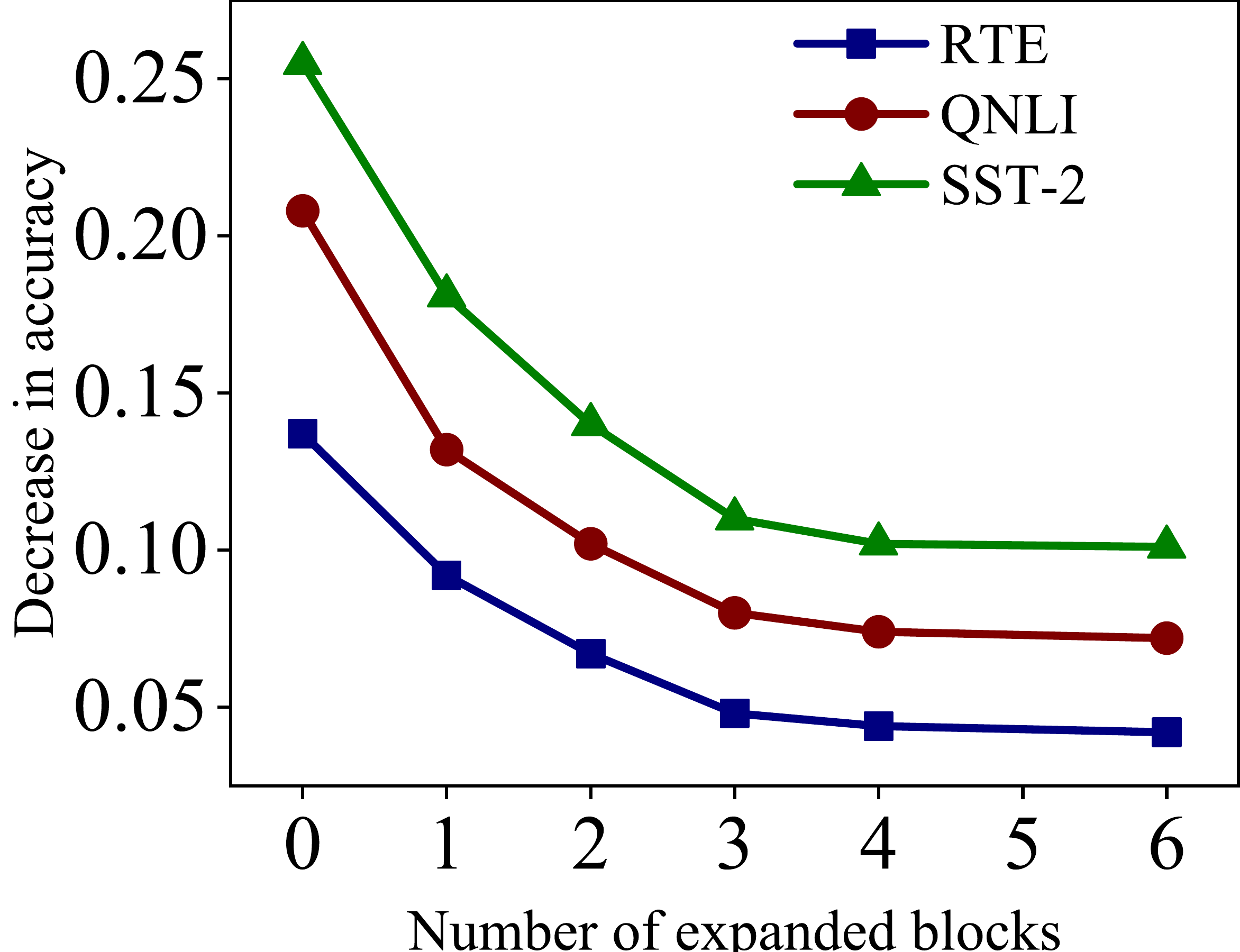}
    }
    \caption{Impact of expanded block number on downstream task and general task performance.}
    \label{fig:number}
\end{figure}

Given the limitations of existing FedFT methods in terms of structured knowledge isolation and adaptation to heterogeneity, we observe that one of the important reasons for catastrophic forgetting lies is the failure to structurally isolate the general abilities obtained from pre-training and the specific task abilities obtained during fine-tuning \cite{delange2022continual}. As a result, the learning of new tasks inevitably disrupts or overwrites the previously acquired knowledge. Moreover, the inherent data heterogeneity in federated settings further amplifies the issue, as model struggles to accommodate diverse local data distributions across clients, severely impairing the model's generalization ability.

Building upon this insight, we further observe that during the training of deep neural networks, the gradient norms of different layers reflect their relative importance in knowledge transfer and feature extraction \cite{evci2022gradient}. Layers with larger gradient norms typically play a more critical role in information processing and have a greater impact on model performance.  Following the observation, we explore a new strategy that inserts trainable expansion modules after the identified key layers, which are determined by their gradient norms, and freezes the parameters of the original model. Such approach enables a structured separation between general and task-specific knowledge and provides architectural adaptation for downstream tasks.

To refine the details of the expansion strategy, we conduct experiments to evaluate the impact of expanded block quantity and position on the mitigation of catastrophic forgetting and downstream task performance. We use the RoBerta-base model and simulate a federated environment with 10 clients on the MRPC task. After federated fine-tuning, the model is evaluated on general tasks including RTE, QNLI, and SST-2, as well as the downstream task MRPC itself.
As shown in Fig.~\ref{fig:number}, when no expansion modules are inserted, the model suffers the most severe performance degradation on general tasks, showing an accuracy drop of 13.7\% on RTE, 20.8\% on QNLI, and 25.5\% on SST-2, which highlights the severe degree of catastrophic forgetting in federated fine-tuning. As we increase the number of expansion layers, the forgetting is substantially mitigated. Specifically, when three expanded blocks are added, the performance degradation is reduced to 4.8\% on RTE, 8.0\% on QNLI, and 11.0\% on SST-2, offering the most efficient configuration in terms of both accuracy and computational cost. Beyond three layers, the benefits plateau, and the downstream task MRPC improves marginally, reaching 87.0\% accuracy with six expanded blocks. It indicates that simply adding more expanded blocks yields diminishing outcomes.

\begin{figure}[t]
    \centering
    \subcaptionbox{MRPC accuracy under different expanded block positions
\label{fig:different-environment}}[0.49\columnwidth][c]{%
        \includegraphics[width=\linewidth]{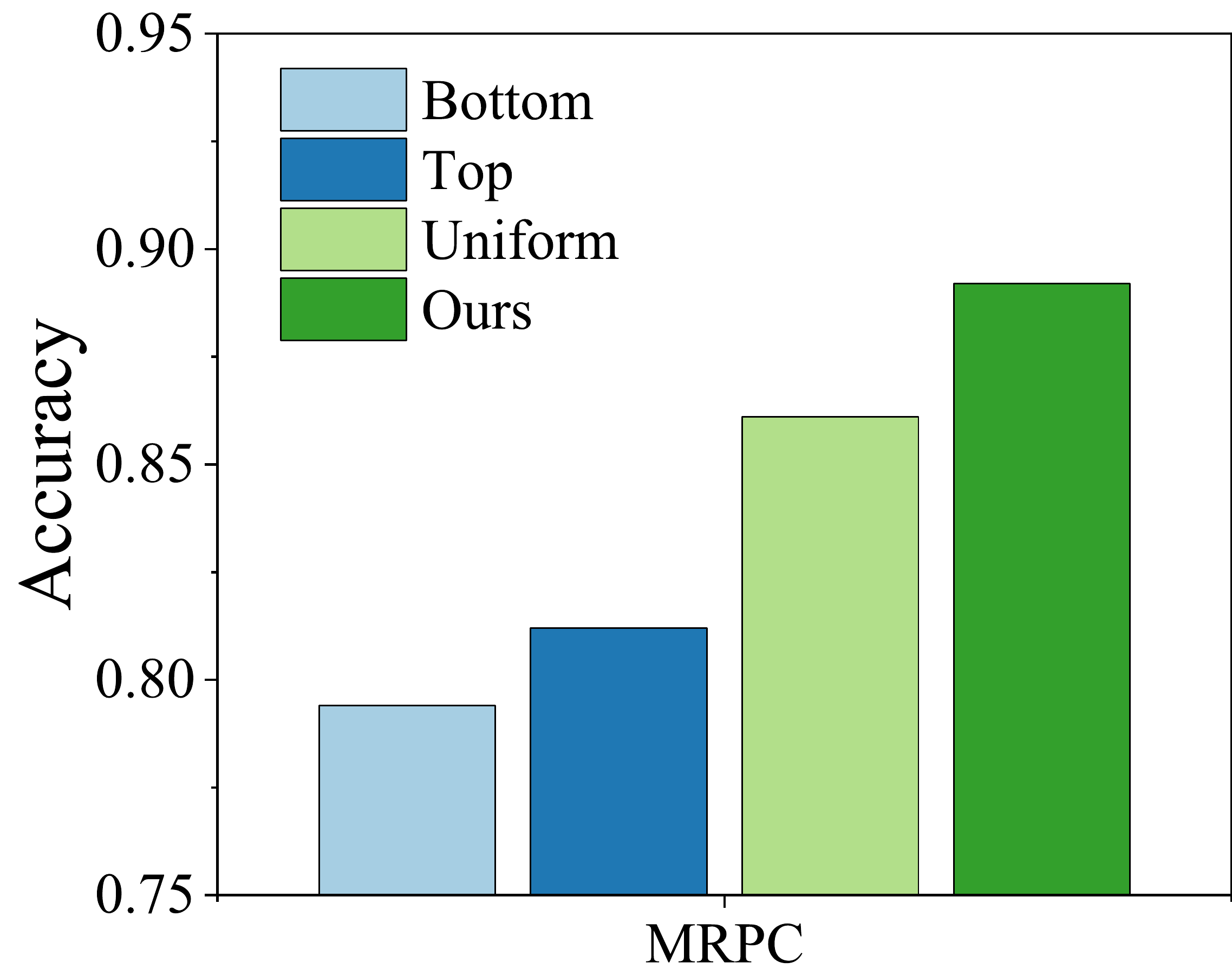}
    }
    \hfill
    \subcaptionbox{Accuracy decrease on RTE, QNLI, and SST-2 under different positions\label{fig:different-method}}[0.49\columnwidth][c]{%
        \includegraphics[width=\linewidth]{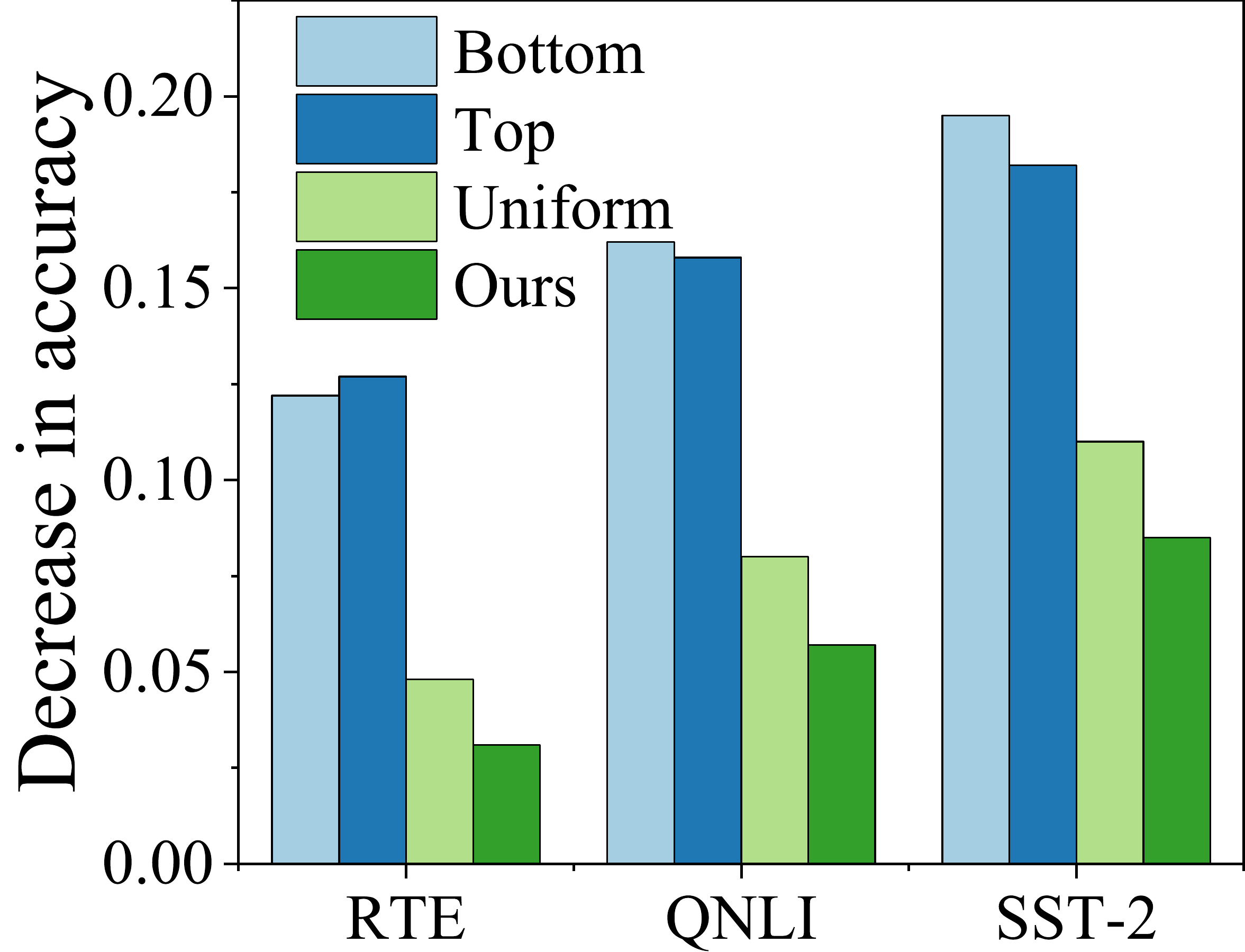}
    }
    \caption{Impact of expanded block placement on downstream task and general task performance.}
    \label{fig:position}
\end{figure}

We further investigate the impact of expansion layer position by fixing the number of expansion layers to three and comparing four position selection strategies: placing all expansion layers near the input side of the model (Bottom), near the output side (Top), uniformly across the model (Uniform), and our proposed method based on gradient norm analysis (Ours).
As shown in Fig.~\ref{fig:position}, the Bottom and Top strategies yield the poorest performance, with MRPC accuracies of 79.4\% and 81.2\%, respectively.
In addition, they lead to substantial accuracy degradation on general tasks—for instance, resulting in 10.2\% and 10.7\% drops on RTE.
The Uniform strategy achieves better generalization and downstream performance, reaching an accuracy of 86.1\% on MRPC and reducing the accuracy drop on RTE to 4.8\%.
In contrast, our gradient-guided approach delivers the best results across all metrics, achieving the highest MRPC accuracy of 89.0\% and further reducing the RTE accuracy drop to only 4.1\%.

These experimental results validate the effectiveness of our design intuition: selectively inserting trainable expanded blocks based on gradient importance enables a structured decoupling of general and task-specific knowledge, significantly mitigating catastrophic forgetting in federated fine-tuning. 
Building on these findings, we next present the comprehensive framework that systematically integrates expanded block design and dynamic training strategies to further enhance model generalization and adaptability in heterogeneous federated environments.



\section{Proposed Framework}\label{sec:framework}
\subsection{Overview of FedBE}

To mitigate catastrophic forgetting and address the challenges of heterogeneity in FedFT, we propose a novel training framework FedBE (FedFT with Block Expansion). The framework integrates gradient-informed block expansion with dynamic trainable blocks allocation, aiming to enhance the adaptability and generalization capability of large models in federated environments.
Specifically, the FedBE framework has two important designs:

(1) Server-Side Block Expansion Strategy:
The server uses a pre-trained model to train on a small-scale proxy dataset (\ding{172}) and monitors the gradient norm at different blocks of the model during the training process (\ding{173}). Based on the monitored gradients, the server estimates block-wise sensitivity to the task and selects the top-$k$ blocks with the highest gradient magnitudes as candidates for expansion (\ding{174}). Trainable expanded blocks are then inserted after the selected blocks. The blocks are carefully designed to minimize disruption to the original model behavior while introducing additional representational capacity for downstream client adaptation.

(2) Client-Aware Trainable Blocks Allocation Strategy:
Before conducting model training on the client side, the server dynamically allocates expanded blocks to each client (\ding{176}) based on two factors: (i) the degree of data distribution uniformity, quantified using Dirichlet-based concentration metrics, and (ii) client-specific resource capabilities. Clients with more uniform class distributions are assigned lower-level (general) expansion layers, as these layers tend to capture fine-grained semantic patterns and require sufficient data coverage for effective training. In contrast, clients with highly skewed distributions receive higher-level (task-specific) expansions, which focus more on abstract representations and can adapt to concentrated task semantics. This allocation strategy allows each client to focus on the most relevant semantic level according to its data distribution, enhancing overall training performance. Furthermore, based on training cost metrics, such as computation time and memory consumption, which are collected from the previous round, the server dynamically adjusts the number of assigned expanded blocks for each client. This allocation enhances local training effectiveness while preserving shared global representations.

After local training, clients only upload the parameters of their assigned expanded blocks. The server aggregates the updates using weighted strategies that account for data scale, training contribution, and task criticality (\ding{179}). The base model is preserved to ensure global consistency, while expanded blocks evolve to capture client-specific knowledge without incurring significant model drift.

This framework provides a flexible, efficient, and scalable way to adapt pre-trained foundation models to federated learning scenarios, thereby striking a balance between general knowledge retention and task-specific adaptation in a heterogeneous federated environment.

\begin{figure}[t] \centering
    \includegraphics[width=0.5\textwidth]
    {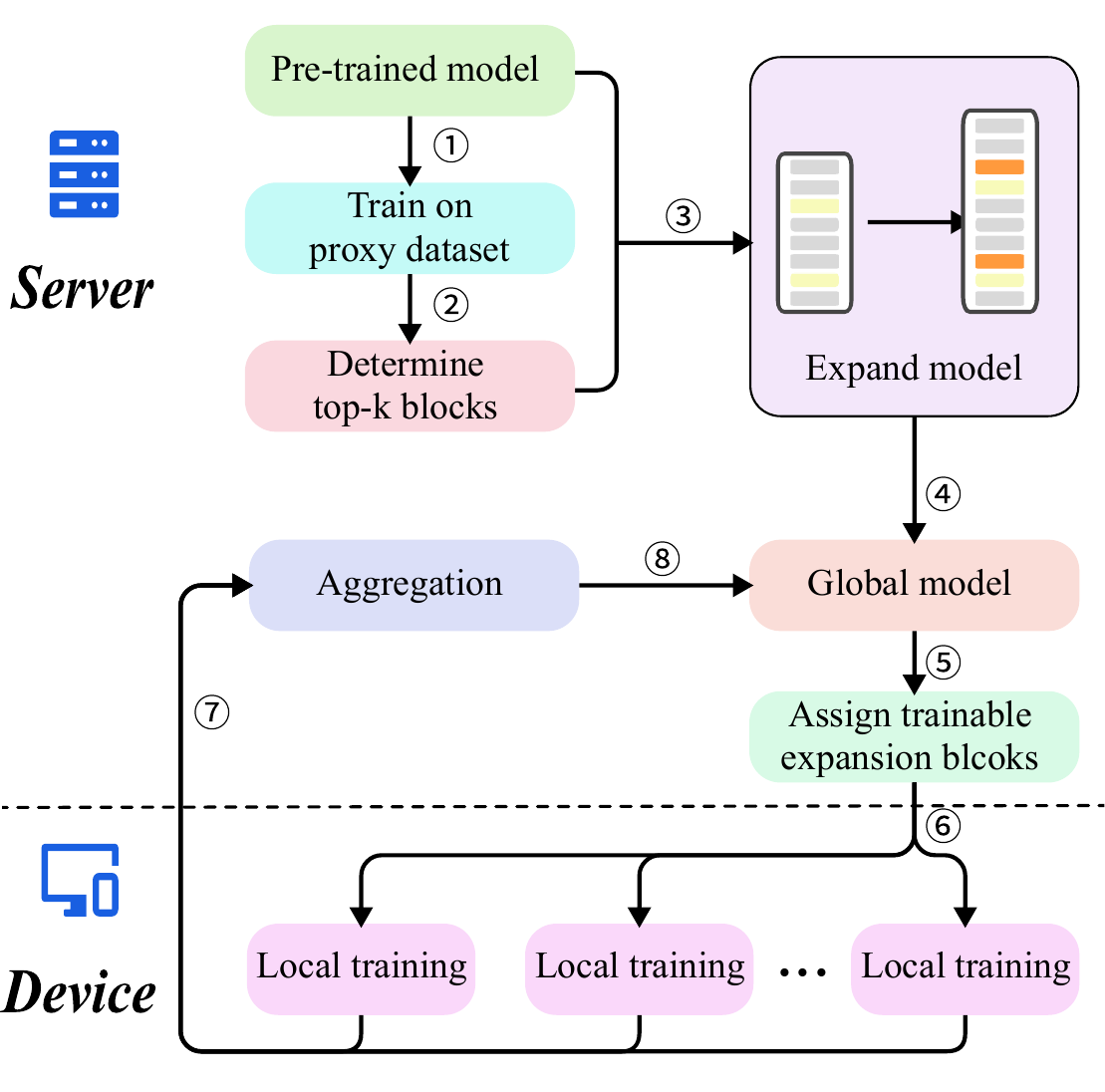}
    \caption{Workflow of FedBE.}\label{fig:flow chart}
\end{figure}

\subsection{Server-Side Block Expansion Strategy}

On the server side, determining the number of expansion layers $k$ serves as the starting point of the framework. Regarding the impact of expanded blocks quantity on fine-tuning effectiveness based on our analysis in Section~2 , we observe a consistent improvement in both downstream task accuracy (e.g., MRPC) and resistance to catastrophic forgetting (e.g., RTE, QNLI, SST-2) as the number of expanded blocks increases. As shown in Fig.~\ref{fig:number}, the trend indicates a generally positive correlation between overall performance and the number of inserted expansion modules within a reasonable range.

To ensure deployment efficiency, we consider both the additional computational cost and the potential inference latency introduced by the expansion layers. Given the system's resource capacity such as memory budget and floating point operations (FLOPs) constraints, we define the number of expansion layers $k$ to satisfy:

\begin{equation}
\Delta P(k) \leq \Delta P_{\text{max}}, \quad \Delta \text{FLOPs}(k) \leq \Delta \text{FLOPs}_{\text{max}}
\end{equation}
where $\Delta P(k)$ and $\Delta \text FLOPs(k)$ denote the additional parameters and computational cost induced by the $k$ expanded blocks, respectively. Solving the constraint yields a suitable value of $k$ that satisfies predefined system-level limits, including memory usage, computational throughput, and inference latency. The strategy provides a practical trade-off between model capacity and computational feasibility, ensuring that the expanded model remains lightweight and deployable across heterogeneous clients.

After selecting the number of expansion layers, the server proceeds to determined the optimal insertion positions. To guide the process, it monitors the gradient norms of each block and computes a composite score $S(l)$ for each layer:

\begin{equation}
    d(l, E_k) = \min_{l' \in E_k} |l - l'| 
\end{equation}
\begin{equation}
    S(l) = \frac{g(l)}{\max_{l'} g(l')} + \lambda \cdot \frac{d(l, E_k)}{L}
\end{equation}
where $g(l) = |\nabla_l|_2$ denotes the $L_2$ norm of the gradient at the $l$-th layer, and $d(l, E_k)$ indicates the distance between the current block and the set of already selected expanded blocks $E_k$.
 The hyperparameter $\lambda$ serves as a balancing factor that controls the influence between gradient magnitude and spatial distribution of expanded blocks. A larger $\lambda$ increases the penalty for selecting layers that are close to each other, thereby promoting a more uniform distribution of expanded blocks throughout the model, but also weakening the influence of gradient norm. Conversely, a smaller $\lambda$ prioritizes layers purely based on their gradient magnitude, potentially resulting in  expanded blocks being clustered together in a few regions. To prevent the expanded blocks from becoming overly concentrated, which could limit the representational diversity and the model’s adaptability across different abstraction levels, we introduce this spatial distribution penalty term. To mitigate the impact of the penalty term on $S(l)$, $\lambda$ is typically set within the range of $[0,1]$.
 The method prioritizes layers with significant gradient contributions while ensuring a reasonable spatial distribution, preventing the expansion layers from being overly concentrated or too sparse. The server ranks all layers based on their composite scores and selects the top-$k$ layers for expansion.

\begin{algorithm}[t]
\caption{Select Expansion Layers}
\label{alg:select_expansion_layers}
\KwIn{
  Average gradient norm for each of the $L$ layers $G = [g_1, g_2, \dots, g_L]$;\\
  $k$: Number of expansion layers;\\
  $\lambda$: Distance regularization coefficient;
}
\KwOut{
  $E_k$: Set of selected layer indices for expansion;
}

Initialize $E_k \leftarrow \emptyset$\;

Normalize gradient scores: $s_\text{grad}(l) \leftarrow g_l / \max(G), \quad \forall l \in \{1, \dots, L\}$\;

\For{$t \leftarrow 1$ \KwTo $k$}{
    Initialize score dictionary $S \leftarrow \emptyset$\;

    \For{each candidate layer $l \in \{1, \dots, L\} \setminus E_k$}{
        \eIf{$E_k$ is empty}{
            $d_\text{penalty}(l) \leftarrow 0$\;
        }{
            $d_\text{penalty}(l) \leftarrow \min_{l' \in E_k} \frac{|l - l'|}{L}$\;
        }
        Compute composite score: $S(l) \leftarrow s_\text{grad}(l) + \lambda \cdot d_\text{penalty}(l)$\;
    }

    Select layer with highest score: $l^* \leftarrow \arg\max_{l \notin E_k} S(l)$\;
    
    Update expansion set: $E_k \leftarrow E_k \cup \{l^*\}$\;
}

\Return $E_k$\;
\end{algorithm}

Once the expansion layers are selected, the implementation involves inserting parameter-initialized expansion modules after the target layers. Let the original model comprise \(L\) layers \(\left( \phi_0, \phi_1, \dots, \phi_L \right)\). For each selected layer \(\phi_l\), we append an expansion layer \(\phi_{\text{expand}_l}\) whose architecture mirrors \(\phi_l\) (e.g., matching dimensions in fully connected layers or normalization modules). Crucially, the parameters of \(\phi_{\text{expand}_l}\) are initialized by copying the parameters from \(\phi_l\) followed by zero-initialization to ensure non-interference with the original model’s behavior at initialization. Specifically:

\begin{itemize}
    \item \textbf{Weighted modules (e.g., fully connected layers):} The weights \(W_{\text{expand}_l}\) and biases \(b_{\text{expand}_l}\) are set to zero, i.e., 
    \[
    W_{\text{expand}_l} \leftarrow 0, \quad b_{\text{expand}_l} \leftarrow 0.
    \]
\end{itemize}

The design guarantees that the expanded model \(\phi_{\text{expanded}}\) initially produces identical outputs to the original model \(\phi\), since \(\phi_{\text{expand}_l}(x) = 0\) for any input \(x\) at initialization. 
For architectures with residual connections (e.g., LLaMA), the expansion layers are seamlessly integrated into the residual pathway. Let the original residual block be
\begin{equation}
x_{l+1} = x_l + \phi_l(x_l).
\end{equation}
After expansion, it becomes:
\begin{equation}
    x_{l+1} = x_l + \phi_l(x_l) + \phi_{\text{expand}_l}(\phi_l(x_l)),
\end{equation}
where the third term on the right-hand side is initialized to zero and activated only during training. It ensures compatibility with architectures relying on skip connections without disrupting their initial behavior. The strategy preserves the pre-trained knowledge in \(\phi\) while allowing the expansion layers to adapt incrementally to heterogeneous client data through FedFT.

\subsection{Client-Aware Trainable Blocks Allocation Strategy}

In federated learning, the significant differences in data distributions and resource capabilities among clients present challenges for training efficiency and overall performance. To address these issues, we propose a dynamic trainable blocks allocation strategy that assigns appropriate training tasks to each client based on its data heterogeneity and resource capability. The strategy aims to improve model adaptation in the presence of client heterogeneity, and to facilitate efficient global optimization.

The data heterogeneity of each client is quantified using the concentration parameter of a Dirichlet distribution. Specifically, assuming the global data is partitioned into \( n \) clients across \( C \) classes, the data distribution for client \( i \) follows a Dirichlet distribution with parameter \( \alpha_i \), where a smaller \( \alpha_i \) indicates higher data heterogeneity (concentration in fewer classes), and a larger \( \alpha_i \) approaches an independent and identically distributed (IID) scenario. To mitigate sensitivity to extreme values, we define the heterogeneity metric as:

\begin{equation}
    D_i = - \log(\alpha_i + \epsilon), \quad \epsilon = 10^{-6}
\end{equation}
Using the data heterogeneity metric, we prioritize the assignment of expanded layers based on their positions. For clients with lower data heterogeneity (larger \(\alpha_i\)), expanded layers closer to the input end are prioritized, as these layers are responsible for extracting general features suitable for uniformly distributed data. Conversely, for clients with higher data heterogeneity (smaller \(\alpha_i\)), expanded layers closer to the output end are prioritized to focus on learning abstract and specific features from their concentrated data distribution. The priority score for layer \( l \) on client \( i \) integrates normalized metrics:

\begin{equation}
    T(l) = w_d \cdot \frac{D_i}{\max_j D_j} + w_t \cdot \frac{T_{i,l}}{1 + \sum_{l'} T_{i,l'}} + w_r \cdot \frac{R_i}{\max_j R_j}
\end{equation}
where \( T_{i,l} \) is the historical training count of layer \( l \) by client \( i \) (weighted inversely to balance under-trained layers), and \( R_i \) denotes the resource capability score (computational power, memory). 
The weights $w_d$, $w_t$, and $w_r$, are predefined to balance the influence of data distribution, training counts, and resource availability. Specifically, the coefficient $w_d$ controls the emphasis on client data heterogeneity, $w_t$ balances the historical training counts, and $w_r$, accounts for the clients' resource constraints (e.g. computational power,memory availability). Typically, these weights satisfy $w_d$ + $w_t$ + $w_r$ = 1.

The server ranks all expanded layers based on their priority scores \( T(l) \) and selects the top-\( k \) layers as the task set for each client, defined as:

\begin{equation}
    E_i(t) = \{ l \mid l \in \text{top-}k \text{ layers by } T(l) \}
\end{equation}
In each training round, the server dynamically adjusts the expanded layer set \( E_i \) for each client to match its resource capability and data distribution.

Additionally, we design a dynamic trainable blocks size adjustment mechanism to address resource heterogeneity among clients. In the initial stage, all clients are assigned the same number of expanded layers for training. However, in subsequent rounds, the server adjusts the trainable blocks size based on feedback from each client, such as training time and memory usage. The task size adjustment formula is given by:

\begin{equation}
    |E_i^{(t+1)}| = \min\left(\max\left(\frac{\tau}{T_i}, 1\right), k\right)
\end{equation}
where \( |E_i^{(t+1)}| \) is the number of expanded layers assigned to client \( i \) in the next round, \(\tau\) represents the global target training time, \( T_i \) is the actual training time for client \( i \) in the previous round, and \( k \) is the total number of expanded layers.

Through the dynamic task allocation strategy, the server assigns appropriate expanded layers to clients based on their resource capabilities and data distribution characteristics. This approach not only improves training efficiency but also significantly mitigates the negative impact of data heterogeneity on global model performance. By comprehensively considering data heterogeneity, resource constraints, the proposed strategy enhances the model’s generalization ability and convergence stability while preserving data privacy in federated learning scenarios.




\section{Performance Evaluation}\label{sec:evaluation}
\subsection{Experimental Setup}

\begin{table}[t]
\centering
\caption{Technical specifications of Jetson kits.}
\setlength{\textfloatsep}{6pt}

\vspace{0.2cm}

\begin{tabular}{|p{2cm}|p{3.2cm}|p{2.5cm}|}
\hline
\textbf{Jetson Kit} & \textbf{AI Performance} & \textbf{ROM} \\ \hline
Jetson TX2 & 1.33 TFLOPS & 8 GB LPDDR4 \\ \hline
Jetson NX & 21 TOPS & 8 GB LPDDR4x \\ \hline
Jetson AGX & 22 TOPS & 32 GB LPDDR4x \\ \hline
\end{tabular}

\vspace{0.4cm}

\begin{tabular}{|p{2cm}|p{3.2cm}|p{2.5cm}|}
\hline
\textbf{Jetson Kit} & \textbf{GPU Type} & \textbf{GPU Frequency} \\ \hline
Jetson TX2 & 256-core Pascal & 0.85 GHz \\ \hline
Jetson NX & 384-core Volta & 0.8 GHz \\ \hline
Jetson AGX & 512-core Volta & 0.9 GHz \\ \hline
\end{tabular}

\vspace{0.4cm}

\begin{tabular}{|p{2cm}|p{3.2cm}|p{2.5cm}|}
\hline
\textbf{Jetson Kit} & \textbf{CPU Type} & \textbf{CPU Frequency} \\ \hline
Jetson TX2 & Denver 2 and ARM 4 & 1.2 GHz \\ \hline
Jetson NX & 6-core Carmel ARM 8 & 1.2 GHz \\ \hline
Jetson AGX & 8-core Carmel ARM 8 & 1.45 GHz \\ \hline
\end{tabular}

\end{table}

\textbf{System Implementation.} To validate the performance of FedBE in a simulated scenario, we conduct extensive experiments on a prototype system. The system consists of a parameter server (PS) and 80 clients. Specifically, the PS runs on a high-performance workstation equipped with an Intel(R) Xeon(R) Platinum 8358P CPU (2.60GHz, 128 cores), 8 NVIDIA RTX A6000 GPUs (48GB memory per GPU), and 512GB of RAM. In addition, we configure 80 NVIDIA commercial development kits, including 30 Jetson TX2, 40 Jetson NX, and 10 Jetson AGX kits, forming a heterogeneous device system. The technical specifications for each Jetson kit are shown in Table 1.





The software platform is built on Docker Swarm \cite{merkel2014docker}, a distributed software toolkit designed for efficiently building and managing distributed systems while providing real-time monitoring of the operational status of each device. We use PyTorch as the deep learning framework to support model training across different devices \cite{paszke2019pytorch}. Moreover, we employ MPI (Message Passing Interface) \cite{gropp1999using}, a set of functions used to simplify communication between the PS and the devices.

\textbf{System Heterogeneity Setup.} To reflect the inherent computational differences among devices, we deploy various Jetson platforms in the prototype system, including Jetson TX2, NX, and AGX. Each device supports different operating modes, allowing it to adjust its computing power as needed. Jetson TX2 provides four different modes for configuration, while Jetson NX and AGX offer up to eight modes, covering a wide range from low to high performance. These modes provide flexibility to dynamically adjust computing resources according to actual requirements. For instance, the Jetson AGX, in its highest-performance mode (mode 0), is up to 100 times faster than Jetson TX2 in its lowest-performance mode (mode 1). This variance not only significantly affects model training efficiency but also reflects the large computational differences between devices.

Additionally, considering the fluctuation of device computing resources during long-term operation, we set up a mechanism to randomly switch the running mode every 20 rounds. The setup helps simulate performance fluctuations that devices may face at different times, ensuring that the experimental results more accurately reflect the instability of device performance in real-world applications.

\textbf{Applications and Models.} We evaluate the performance of the system on two typical NLP applications:

\begin{itemize}
    \item \textbf{Sentiment Analysis}: Sentiment analysis is an important task in natural language processing (NLP) that aims to identify and extract sentiment information from text, typically determining whether the text expresses a positive, negative, or neutral sentiment. In this study, we use the \textbf{IMDB} dataset, a widely used dataset for sentiment analysis, containing 50,000 movie review texts, with 25,000 for training and 25,000 for testing. Each review is labeled as positive or negative sentiment. The goal of the sentiment analysis task is to classify the sentiment of each review accurately, providing data support for applications such as movie, product, or service recommendations.
    \item \textbf{Text Classification}: Text classification is another fundamental task in NLP, aiming to categorize text into predefined categories based on its content. We use the \textbf{AG News} dataset for text classification experiments. This dataset is a multi-class news classification task, containing 120,000 news articles classified into four major categories: world, sports, business, and science. Each news article is clearly labeled with a category, with the training set containing 120,000 samples and the test set containing 7,600 samples. This dataset is widely used for news classification, information retrieval, and similar tasks. Text classification can help automatically analyze large volumes of text data from news, social media, forums, and more. In this task, the model's goal is to extract meaningful features from the articles and classify them into the correct categories.
\end{itemize}

For evaluating the system's adaptability to models with different parameter scales, we choose the following two models for evaluation:
\begin{itemize}
    \item \textbf{RoBerta-base} is a transformer-based pre-trained language model with 125 million parameters, comprising 12 transformer blocks, each with 768 hidden units and 12 self-attention heads. It improves on BERT by removing certain limitations, using larger datasets, and training for a longer duration, thereby enhancing performance on a variety of NLP tasks.
    \item \textbf{BERT-large} is a larger version of BERT with 340 million parameters, consisting of 24 transformer blocks, each with 1024 hidden units and 16 self-attention heads. Compared to \texttt{BERT-base}, \texttt{BERT-large} has stronger learning capacity, suitable for tasks requiring high precision, but it comes with larger computational and memory costs.
\end{itemize}

\textbf{Baseline Methods.} To comprehensively evaluate the effectiveness of our method, we compare it against the following three baseline methods:
\begin{itemize}
    \item \textbf{FedNLP} \cite{lin2021fednlp}: A FedFT framework designed for distributed training across multiple devices. In this method, all devices use the same global model, fine-tune the model on local data, and periodically upload the updated parameters to the server for aggregation. FedNLP allows large-scale model fine-tuning while preserving privacy.
    \item \textbf{FedLoRA} \cite{zhang2023fedpetuning}: A strategy that introduces LoRA module into FedFT. All devices use the same LoRA rank to fine-tune the model locally, updating only part of the model's weights through low-rank matrix approximations.
    \item \textbf{FedAdapters} \cite{cai2022fedadapter}: An adapter-based fine-tuning method that optimizes federated learning for device heterogeneity. In this approach, each device fine-tunes only the adapter rather than the entire model. Additionally, a dynamic search mechanism is introduced to automatically select the optimal adapter configuration based on the device's specific conditions.
\end{itemize}
These baseline methods provide different fine-tuning strategies, including global model updates, low-rank adjustments, and adapter-based fine-tuning, serving as effective benchmarks for comparison with our proposed method.

\textbf{Evaluation Metrics.} To comprehensively assess the performance of FedBE and the baseline methods, we use the following evaluation metrics:

\begin{itemize}
    \item \textbf{Test Accuracy}, which  is the proportion of correctly predicted samples, evaluates the performance of models trained by different methods on the test set. We record the accuracy of the global model (aggregated model) on the test data after each training round to compare the final performance across methods.
    \item \textbf{Training Time-Accuracy} measures the total training time required to achieve the target accuracy. To ensure a fair comparison, we set the target accuracy to the lowest accuracy among the four methods. We record and accumulate the training time per round and compute the average waiting time for each method to reflect differences in training efficiency.
    \item \textbf{Catastrophic Forgetting Degree} measures the extent of catastrophic forgetting, where the model's accuracy on the benchmark test set may significantly decrease during training. By comparing the model's accuracy on the test set before and after training, we can assess the model's stability and effectiveness over the course of long-term training.

\end{itemize}

\textbf{Experimental Parameters.} By default, each experiment runs for 50 rounds, with the learning rate set to 2e-5. The batch size is 32, and the maximum sequence length for input texts is set to 256. For the hyperparameters mentioned in Chapter 3, we set $\lambda$ = 0.5 (equation 5) to balance the influence of gradient norm and expanded block distribution density on the expansion position, and set $w_d$ = 0.5, $w_t$ = 0.3, $w_r$ = 0.2 (equation 9) to effectively balance the effects of client data heterogeneity, training history, and resource constraints on training block selection.

\begin{table}[t]
\centering
\caption{Performance comparison on AG News and IMDB datasets.}
\normalsize 
\begin{tabular}{|p{1.8cm}|c|c|c|c|}
\hline
\textbf{Model} & \multicolumn{2}{c|}{\textbf{RoBerta-base}} & \multicolumn{2}{c|}{\textbf{BERT-large}} \\
\hline
\textbf{Dataset} & AG News & IMDB & AG News & IMDB \\
\hline
\textbf{FedBE} & 93.2\% & 93.6\% & 87.5\% & 82.2\% \\
\hline
\textbf{fedNLP} & 91.2\% & 91.9\% & 86.5\% & 81.4\% \\
\hline
\textbf{fedAdapter} & 91.6\% & 91.6\% & 85.8\% & 81.1\% \\
\hline
\textbf{fedLoRA} & 90.3\% & 91.4\% & 84.4\% & 80.2\% \\
\hline
\end{tabular}
\vspace{0.5cm}
\label{table:1}
\end{table}

\subsection{Experimental Results}

\textbf{Accuracy Performance:}
We compare the performance of existing methods with FedBE on two NLP tasks, analyzing the models' test accuracy. Table~\ref{table:1} shows the final convergence accuracy of each method on the AG News and IMDB tasks, as well as when applied to the RoBerta-base and BERT-large models.

\begin{table*}[t]
\centering
\caption{Comparison of accuracy on various datasets.}
\normalsize
\setlength{\tabcolsep}{10pt}
\begin{tabular}{p{2.3cm}p{1.3cm}ccccc}
\toprule
\textbf{Model} & \textbf{Dataset} & \textbf{FedBE} & \textbf{FedNLP} & \textbf{FedAdapter} & \textbf{FedLoRA} & \textbf{ORI} \\
\midrule
\multirow{6}{*}{RoBerta-base} 
& MRPC  & 69.52\% & 66.40\% & 63.80\% & 63.20\% & 67.60\% \\
& CoLA  & 71.00\% & 67.00\% & 67.60\% & 68.40\% & 70.92\% \\
& QNLI  & 63.64\% & 60.60\% & 52.12\% & 51.92\% & 52.88\% \\
& QQP   & 69.48\% & 66.20\% & 64.42\% & 63.72\% & 63.72\% \\
& RTE   & 59.30\% & 58.01\% & 53.16\% & 51.76\% & 53.95\% \\
& SST-2 & 87.24\% & 77.40\% & 54.40\% & 54.40\% & 57.44\% \\
\midrule
\multirow{6}{*}{BERT-large} 
& MRPC  & 69.56\% & 68.76\% & 66.74\% & 69.36\% & 74.40\% \\
& CoLA  & 72.42\% & 66.96\% & 68.44\% & 67.24\% & 71.22\% \\
& QNLI  & 58.32\% & 53.04\% & 54.42\% & 54.16\% & 57.68\% \\
& QQP   & 70.16\% & 65.92\% & 66.80\% & 65.60\% & 64.54\% \\
& RTE   & 57.65\% & 51.53\% & 53.78\% & 54.98\% & 58.05\% \\
& SST-2 & 77.96\% & 70.96\% & 62.48\% & 68.28\% & 64.84\% \\
\bottomrule
\end{tabular}
\vspace{0.5cm}
\label{table:2}
\end{table*}

\begin{itemize}
    \item On the AG\_News task, with the RoBerta-base model, our method, FedBE, achieves the highest accuracy (93.2\%), which is an improvement of about 2.0\% to 3.2\% over the existing methods: FedNLP (91.2\%), FedAdapter (91.6\%), and FedLoRA (90.3\%). Specifically, FedBE is 2.0\% higher than FedNLP, 1.6\% higher than FedAdapter, and 3.2\% higher than FedLoRA. This significant performance improvement indicates that FedBE is better at capturing the semantic features of the text when handling news classification tasks, thereby improving classification accuracy. In the BERT-large model, FedBE also achieved an accuracy of 87.5\%, which is 1\% higher than FedNLP, and 1.7\% and 3.1\% higher than FedAdapter and FedLoRA, respectively.
  \item On the IMDB task, FedBE also achieves the best performance on both the RoBerta-base and BERT-large models. For RoBerta-base, FedBE achieves an accuracy of 93.6\%, which is 1.7\% higher than FedNLP (91.9\%), 2.0\% higher than FedAdapter (91.6\%), and 2.2\% higher than FedLoRA (91.4\%). This improvement is particularly important in sentiment analysis tasks, as sentiment classification requires a high level of semantic understanding. FedBE, through its optimized training mechanism, is better able to capture the sentiment tendencies in the text. For BERT-large, FedBE achieved an accuracy of 82.2\%, outperforming FedNLP (81.4\%) and FedAdapter (81.1\%), and is 2\% higher than FedLoRA (80.2\%).
\end{itemize}

Overall, FedBE significantly outperforms other methods on both NLP tasks, demonstrating its superiority in final accuracy. This performance gain stems from FedBE’s core design: the gradient-informed block expansion mechanism, which selectively expand additional transformer blocks based on block-wise gradient signals, allows the model to focus capacity on blocks most sensitive to local task distributions. Moreover, the dynamic trainable blocks allocation strategy at the server side helps to assign different trainable blocks to heterogeneous client data distributions, enhancing adaptation without compromising generalization. These two innovations jointly address the challenges of non-IID data and catastrophic forgetting in federated fine-tuning. Notably, FedBE maintains strong performance across both compact (RoBerta-base) and large-scale (BERT-large) models, validating its scalability and broad applicability.

\textbf{Evaluation of Knowledge Retention:}
Subsequently, we further test the general capabilities of the models fine-tuned with different tasks to verify the advantages of our method in mitigating catastrophic forgetting. Table~\ref{table:2} shows the performance comparison among federated fine-tuning methods (FedBE, FedNLP, FedAdapter, FedLoRA) based on the RoBerta-base and BERT-large models, and the original pretrained models (ORI) on several GLUE benchmark tasks (including MRPC, CoLA, QNLI, QQP, RTE, and SST-2). The test results show that the FedBE method demonstrates significant and consistent advantages in terms of catastrophic forgetting, especially excelling in complex semantic understanding tasks.

In the QNLI task, FedBE achieves an accuracy of 63.64\%, a 10.76\% improvement over the original model (52.88\%), and significantly outperforms FedNLP (60.60\%) and FedAdapter (52.12\%). The remarkable performance improvement indicates that FedBE effectively alleviates forgetting in question-answer reasoning tasks through its block expansion and block selection mechanism. Specifically, FedBE focuses on selecting key blocks for expansion during training, while preserving original parameters to avoid overwriting crucial knowledge, thus maintaining high accuracy in question-answering tasks.
In the SST-2 sentiment analysis task, FedBE reached an accuracy of 87.24\%, a 51.80\% improvement over the original model (57.44\%), and far surpassed FedNLP (77.40\%) and FedAdapter (54.40\%). This result suggests that FedBE balances new task adaptation with historical knowledge retention in sentiment classification tasks.
In both the MRPC and CoLA tasks, FedBE also showed superior performance. In MRPC, FedBE achieves an accuracy of 69.56\%, higher than the original model (67.60\%), and led in all methods. In CoLA, FedBE reaches 72.42\%, a significant improvement over the original model (70.92\%), and also outperforms FedAdapter (68.44\%) and FedNLP (66.96\%). These results demonstrate that FedBE shows stable resistance to forgetting in tasks involving semantic equivalence judgment and grammatical correctness evaluation.
Furthermore, in the QQP task, FedBE achieves an accuracy of 70.16\%, a 6.00 percentage point improvement over the original model (63.72\%), and shows the most consistent performance across all methods. This result indicates that FedBE also has a significant advantage in question similarity matching tasks.

Overall, FedBE significantly outperforms other federated fine-tuning methods in terms of resistance to forgetting on the GLUE benchmarks, highlighting its substantial advantage in mitigating catastrophic forgetting. Through important block selection, block expansion, and dynamic allocation of training blocks to clients, FedBE effectively balances new task adaptation with historical knowledge retention. These mechanisms allow FedBE to capture new data features and improve model performance while preventing the overwriting of existing knowledge, making it an ideal choice for multi-task fine-tuning in federated learning scenarios, especially in applications that require a balance between new task adaptability and original capability retention.

\begin{figure}[t]
    \centering
    \begin{subfigure}[b]{0.5\columnwidth}
        \centering
        \includegraphics[width=\textwidth]{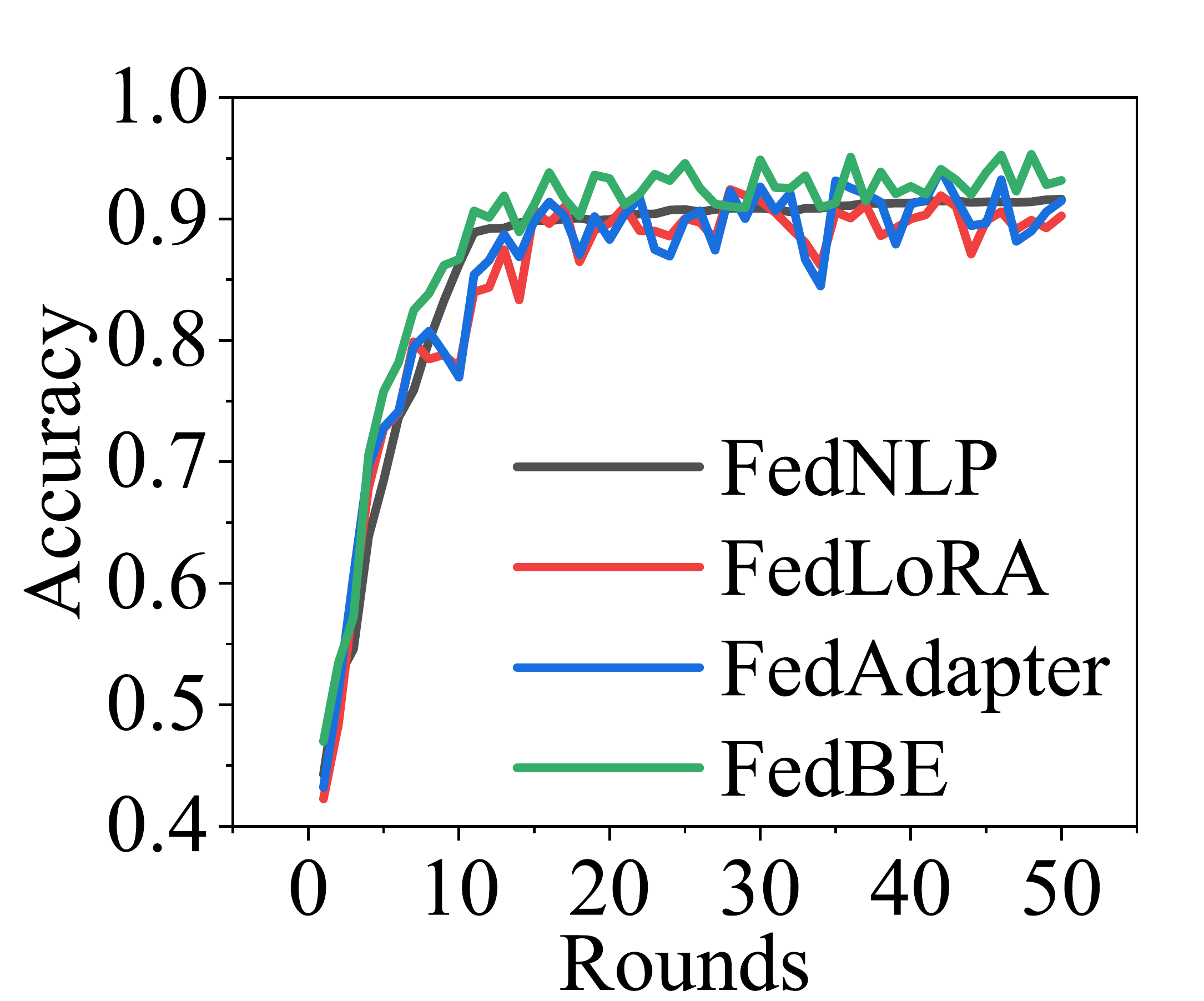}
        \caption{AG News}
    \end{subfigure}%
    \hfill
    \begin{subfigure}[b]{0.5\columnwidth}
        \centering
        \includegraphics[width=\textwidth]{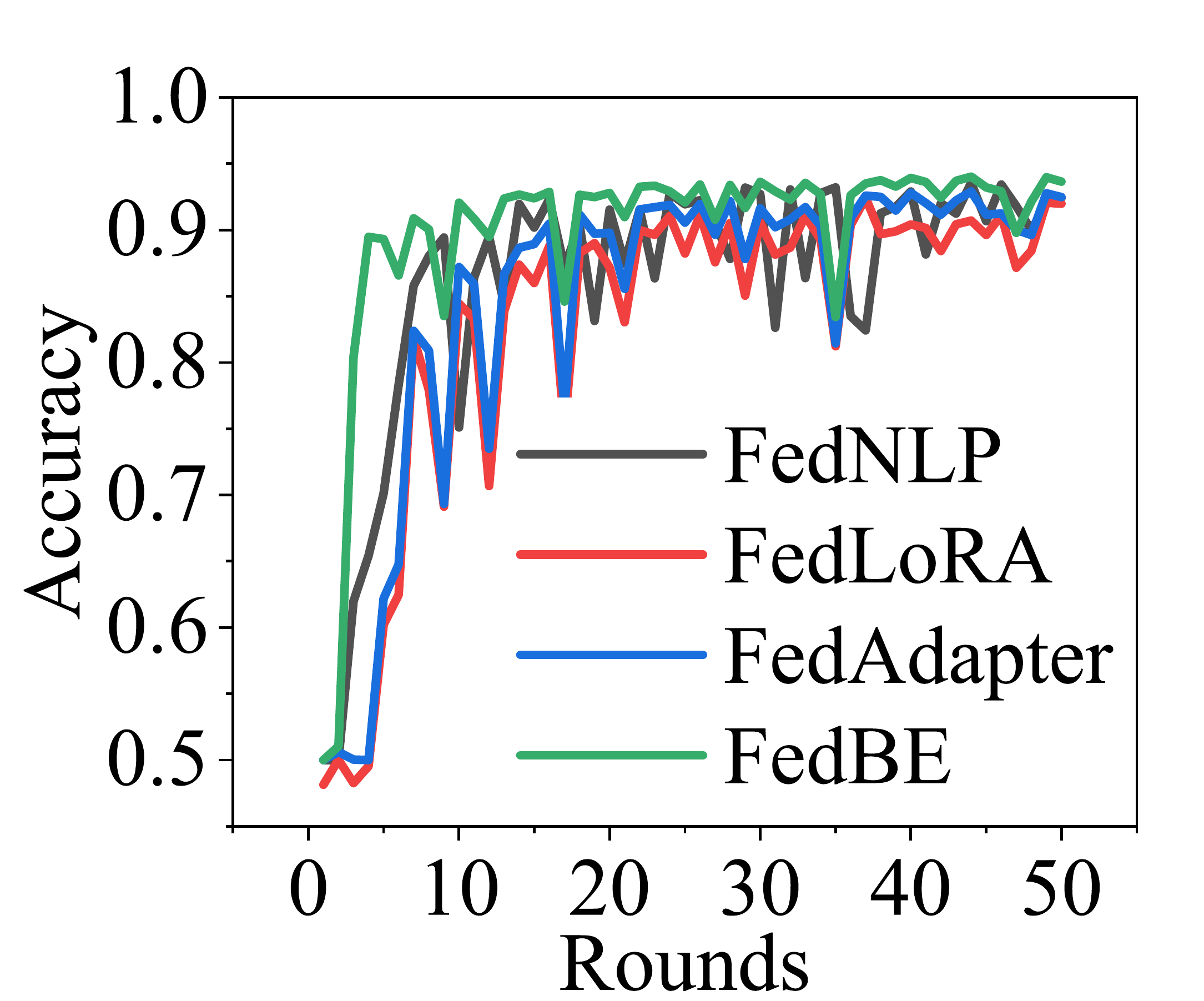}
        \caption{IMDB}
    \end{subfigure}
    
    \caption{Accuracy trends of different methods on AG News and IMDB.}
    \label{fig:line}
\end{figure}
\begin{figure}[t]
    \centering
    \begin{subfigure}[b]{0.5\columnwidth}
        \centering
        \includegraphics[width=\textwidth]{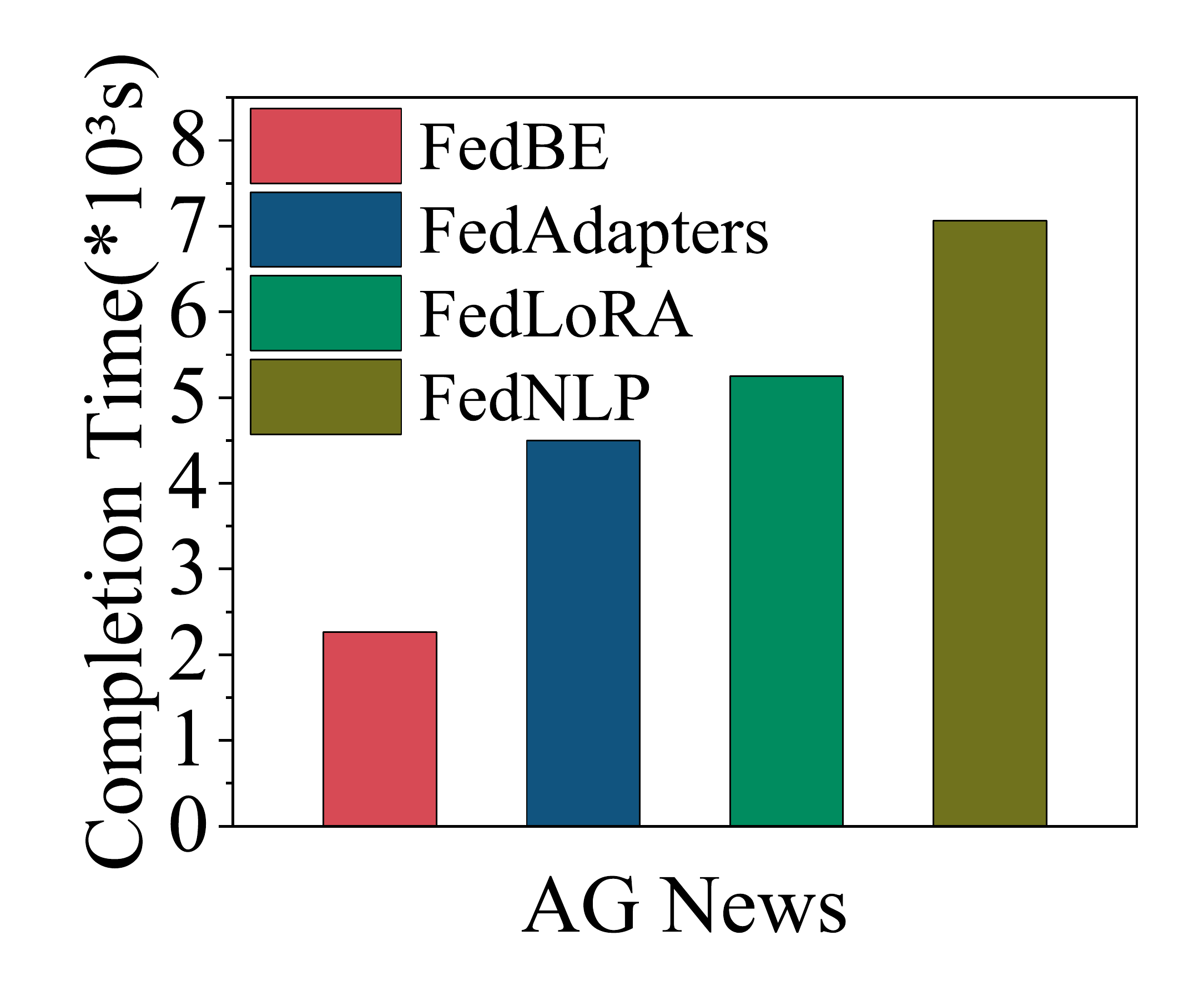}
        \caption{AG News}
    \end{subfigure}%
    \hfill
    \begin{subfigure}[b]{0.5\columnwidth}
        \centering
        \includegraphics[width=\textwidth]{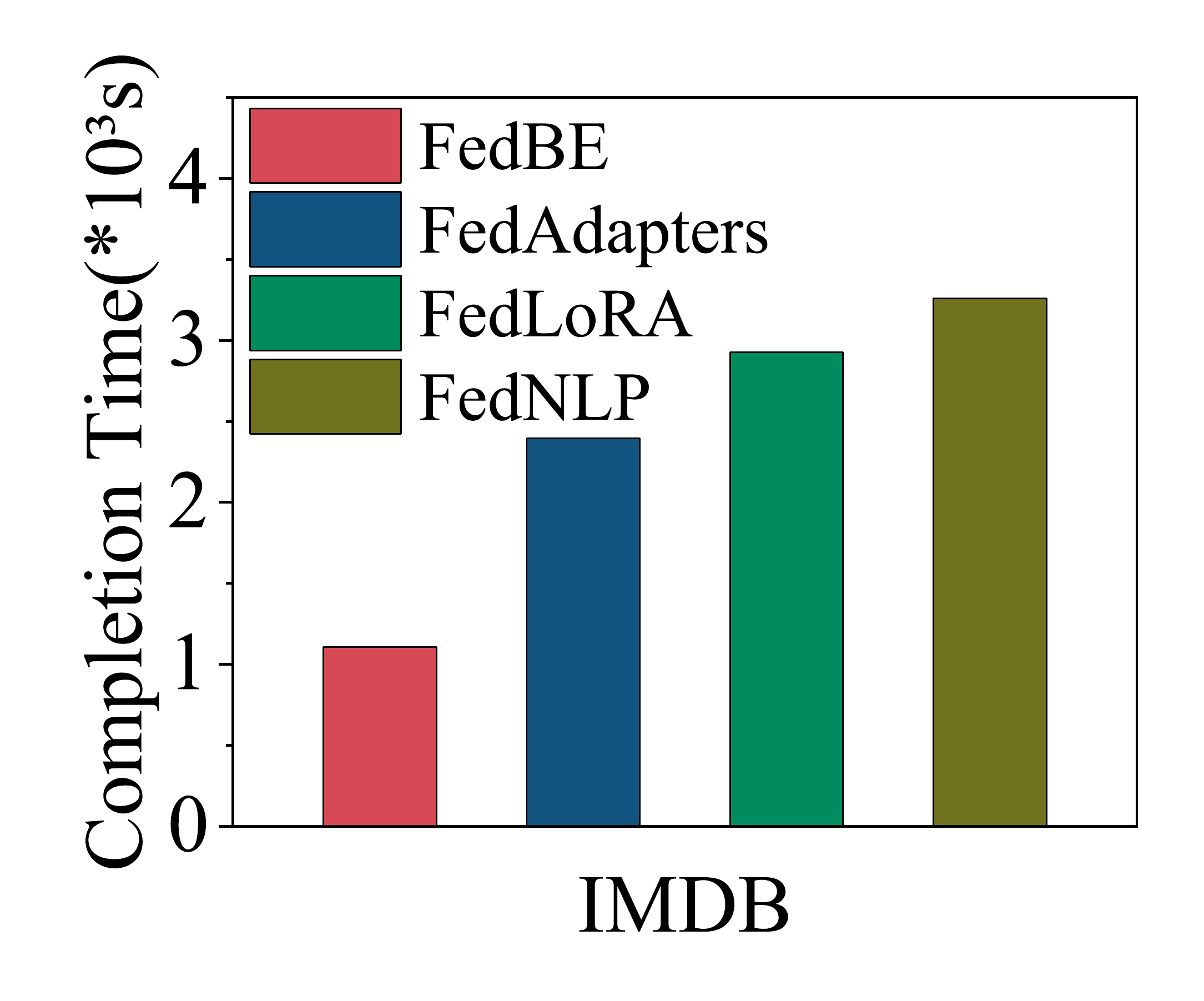}
        \caption{IMDB}
    \end{subfigure}

    \caption{Comparison of training time across different methods on AG News and IMDB.}
    \label{fig:time}
\end{figure}

\textbf{Convergence Speed:}
As shown in Fig.~\ref{fig:line}(a), on the AG News dataset, all four methods achieve high validation accuracy after approximately 20 training rounds. However, FedBE (green curve) rapidly approaches the optimal level within the first 10 rounds and maintains relatively stable fluctuations in the subsequent training phase, demonstrating good adaptability to heterogeneous data. In contrast, the initial convergence speed of FedNLP, FedLoRA, and FedAdapter is generally slower, and their subsequent fluctuations are more pronounced. Similarly, as seen in Fig.~\ref{fig:line}(b), in the IMDB task, FedBE also approaches optimal accuracy in fewer than 10 training rounds and exhibits continuous stability throughout the training process. Although FedNLP, FedLoRA, and FedAdapter gradually reach high accuracy in later rounds, their initial growth speed and later stability are inferior to FedBE.

Meanwhile, Fig.~\ref{fig:time} further illustrates the training time required for each method to reach 90\% validation accuracy on the AG News and IMDB datasets. Specifically, FedBE only takes 2266 seconds on AG News and 1106 seconds on IMDB, which are several times shorter than FedAdapter, FedLoRA, and FedNLP. More specifically, the required training times for FedAdapter, FedLoRA, and FedNLP on the IMDB task are 2394 seconds, 2926 seconds, and 3259 seconds, respectively, all significantly higher than the 1106 seconds of FedBE. The AG News dataset exhibits similar results.

Considering both convergence speed and curve stability, the design of FedBE in terms of important block selection, block expansion, and dynamic allocation of training blocks enables it to quickly capture new data features and improve model accuracy while effectively retaining prior knowledge. This prevents performance fluctuations and forgetting problems commonly observed in multi-task scenarios, thereby demonstrating significant advantages in training rounds and total training time.

\subsection{Ablation Study}

To further verify the effectiveness of FedBE in key mechanisms, we design a series of ablation experiments focusing on the impact of expanded block position decision and dynamic task allocation mechanisms on model performance. By removing these key mechanisms and comparing model accuracy and convergence speed, we can assess their roles in enhancing model adaptability and training efficiency.

\begin{figure}[t]
    \centering
    \begin{subfigure}[b]{0.5\columnwidth}
        \centering
        \includegraphics[width=\textwidth]{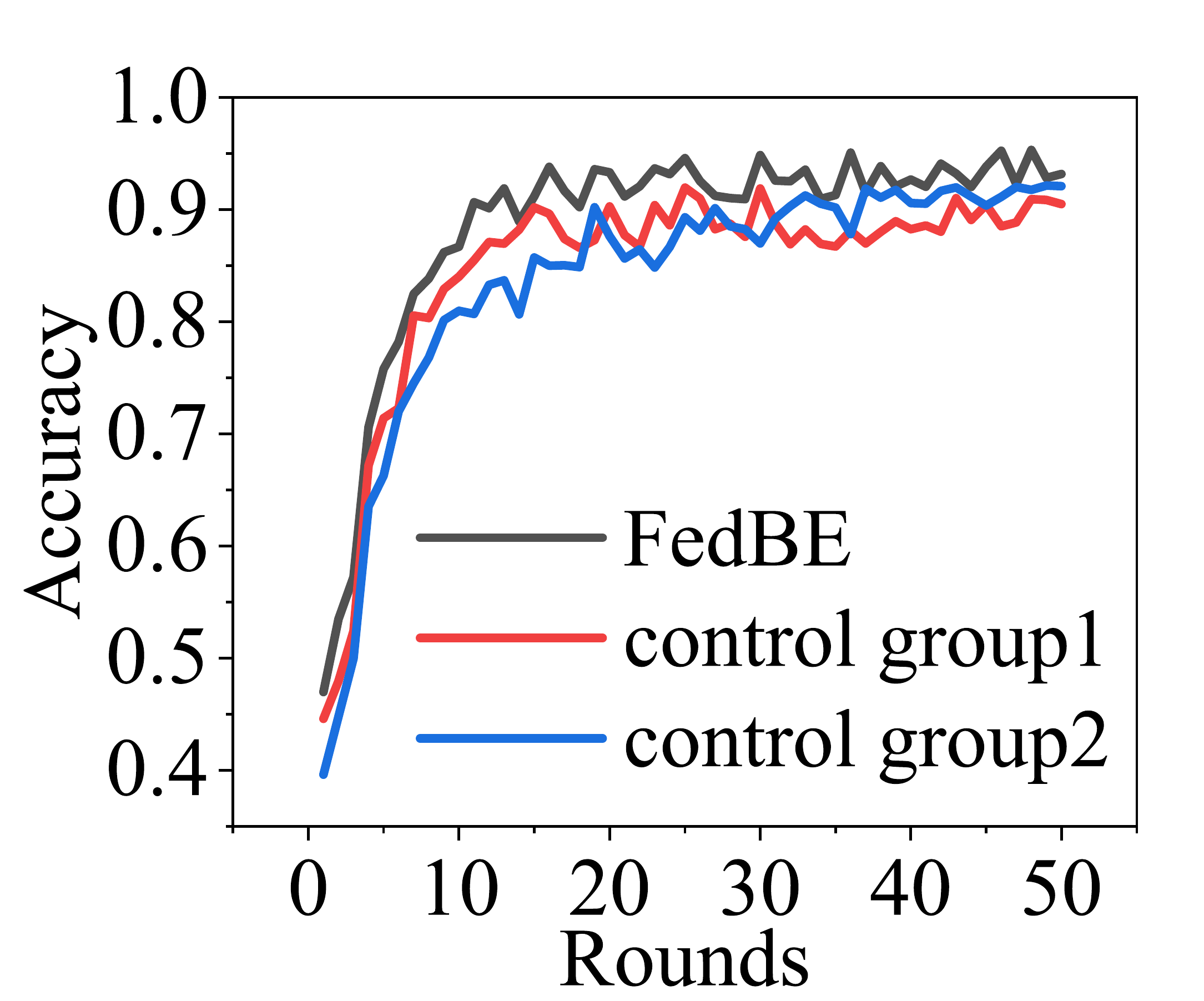}
        \caption{AG News}
    \end{subfigure}%
    \hfill
    \begin{subfigure}[b]{0.5\columnwidth}
        \centering
        \includegraphics[width=\textwidth]{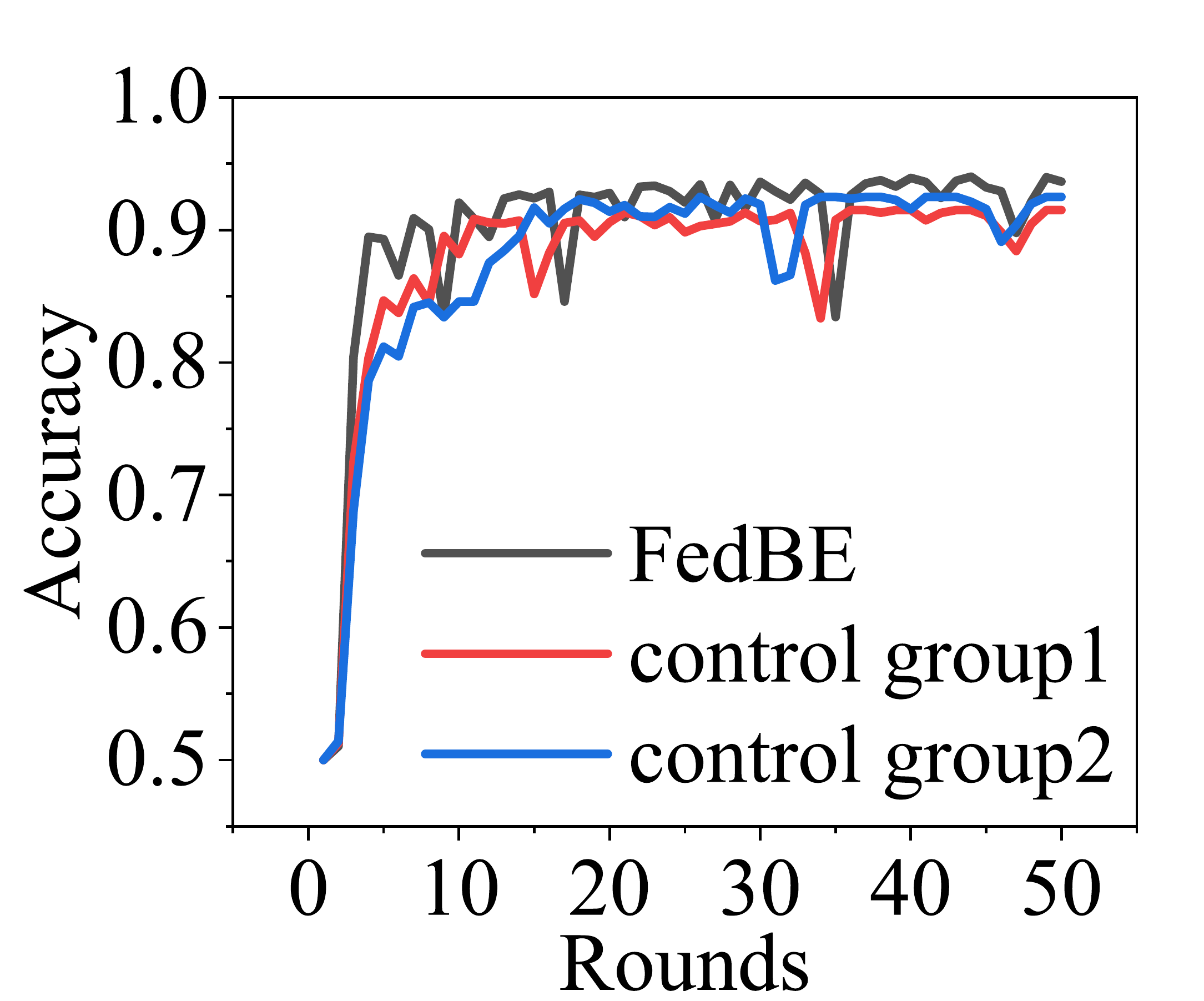}
        \caption{IMDB}
    \end{subfigure}

    \caption{Accuracy comparison under different ablation settings on AG News and IMDB.}
    \label{fig:control_line}
\end{figure}

To analyze the impact of expanded block positioning strategies on model performance and stability, we construct a controlled experiment where the original expanded block positioning strategy of FedBE is removed. Instead, the blocks of the RoBerta-base model are evenly divided into three groups, and an additional expanded block is added at the end of each group for training, while keeping other experimental settings unchanged.
The experimental results, shown as the control group1 curve in Fig.~\ref{fig:control_line}, indicate that on the AG News and IMDB datasets, this strategy leads to a significant drop in overall model accuracy and slower convergence speed compared to the original FedBE. Specifically, as the expanded blocks are not allocated to optimal positions based on task requirements, the model exhibited lower adaptability in capturing new data features and experienced greater performance fluctuations. 
The phenomenon suggests that the selection of expanded block positions is crucial for enhancing the generalization capability of the model. A well-planned expanded block placement can more effectively learn new task information while retaining existing knowledge, thereby facilitating stable optimization in the federated learning environment.
\begin{figure}[t]
    \centering
    \begin{subfigure}[b]{0.5\columnwidth}
        \centering
        \includegraphics[width=\textwidth]{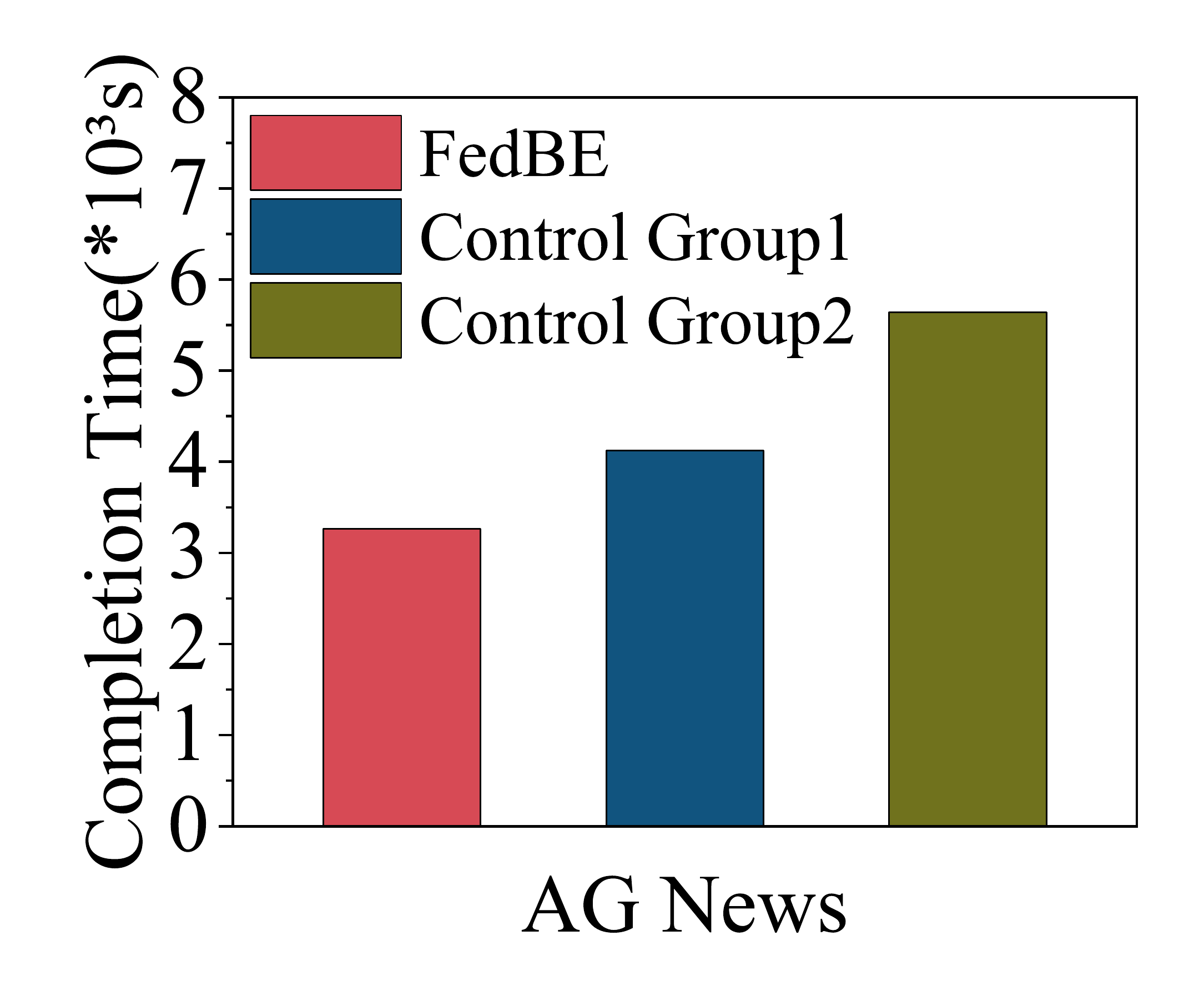}
        \caption{AG News}
    \end{subfigure}%
    \hfill
    \begin{subfigure}[b]{0.5\columnwidth}
        \centering
        \includegraphics[width=\textwidth]{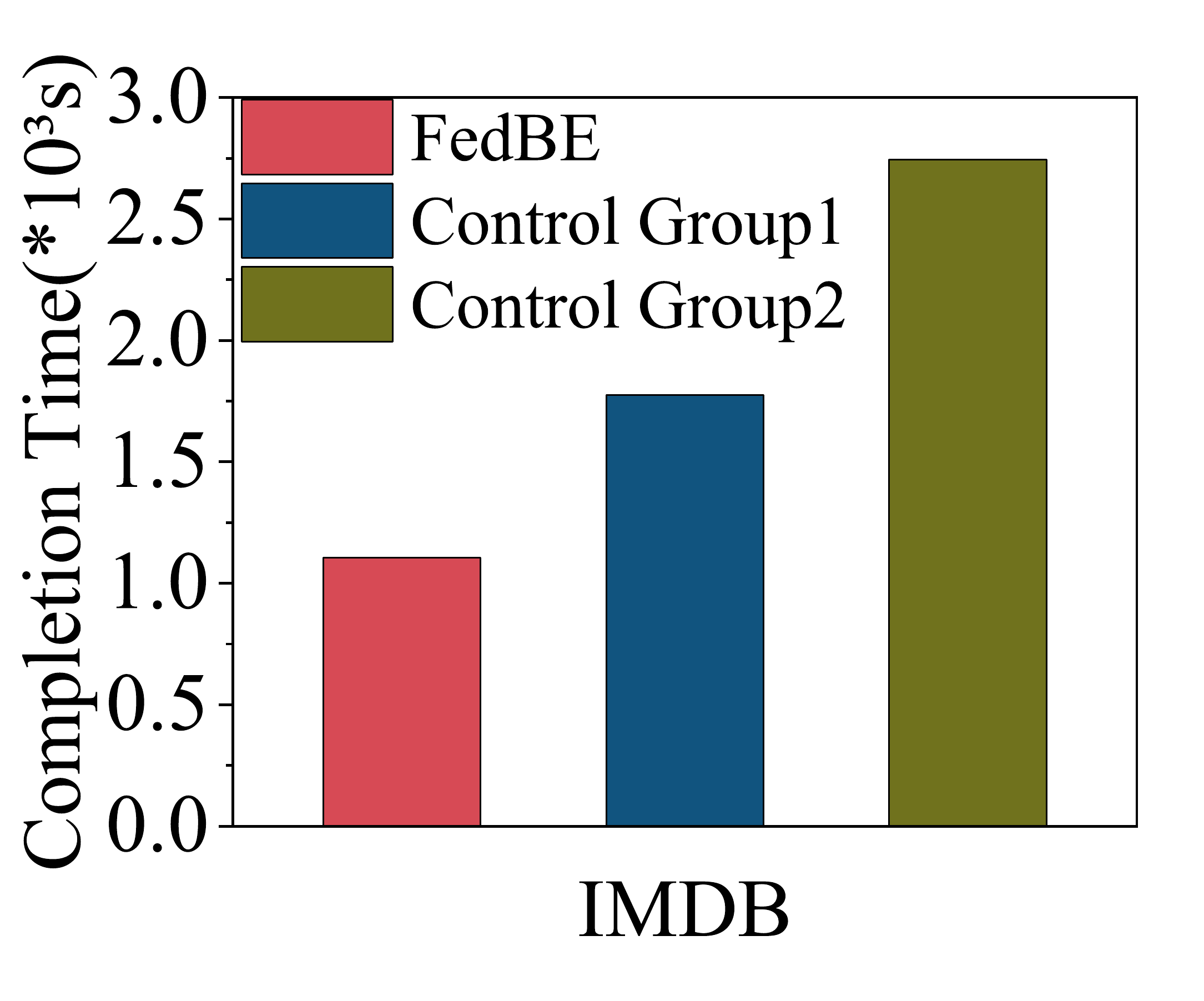}
        \caption{IMDB}
    \end{subfigure}

    \caption{Training time comparison under different ablation settings on AG News and IMDB.}
    \label{fig:control_time}
\end{figure}

In addition, we evaluate the role of the dynamic task allocation mechanism under a heterogeneous federated learning setting. Specifically, we removed FedBE’s dynamic training block selection strategy. In this setting, each client trains all expanded blocks instead of performing personalized training block allocation based on its data heterogeneity and computational resources.
As shown in Fig.~\ref{fig:control_line}, the results for control group2 demonstrate that although the final accuracy difference is not significant compared to the original FedBE, the convergence speed is noticeably slower. On both datasets, the model requires more training rounds to reach the same accuracy. Furthermore, due to the bottleneck effect, devices with limited resources take longer per training round, which makes the difference in training time to reach 90\% accuracy in Fig.~\ref{fig:control_line} more pronounced.
This indicates that the dynamic task allocation mechanism optimizes training according to the heterogeneity of client computing capabilities, mitigating the impact of the bottleneck effect and improving convergence efficiency. Additionally, since not all clients need to train all expanded blocks, computational resources are not excessively consumed on unnecessary block updates, reducing both computational and communication overhead. Therefore, in a federated learning environment, a dynamic block selection strategy is essential for improving training efficiency.

\section{Related Work}\label{sec:related}
Modern natural language processing (NLP) has witnessed rapid advancements, particularly driven by the emergence of transformer based language models, such as BERT \cite{devlin2019bert}, GPT-2 \cite{radford2019language}, and LLaMA \cite{touvron2023llama}. These models have demonstrated great performance across a wide range of tasks, including text classification, machine translation, and question answering \cite{vaswani2017attention}. This success is largely attributed to the pre-training and fine-tuning paradigm \cite{raffel2020exploring}, wherein models are first pre-trained on large scale unsupervised corpora to acquire general purpose language representations, and subsequently fine-tuned on task specific datasets to enable effective adaptation to downstream applications. However, conventional fine-tuning approaches typically rely on centralized data storage and processing, thereby raising significant concerns regarding data privacy, security, and regulatory compliance \cite{regulation2016eu}. In light of the growing emphasis on data protection, federated fine-tuning (FedFT), an extension of federated learning has emerged as a promising alternative, offering wide applicability across various privacy sensitive domains.

Lin et al. were the first to conduct large model fine-tuning in a federated environment by jointly optimizing all model parameters to improve global model performance\cite{lin2021fednlp}. Since then, the rapid advancement of modern fine-tuning techniques has led to the emergence of many parameter efficient fine-tuning (PEFT) methods, such as LoRA \cite{hu2022lora} and Prompt Tuning \cite{lester2021power}. Zhang et al. proposed FedLoRA and validated LoRA's efficiency in the FedFT setting through extensive experiments \cite{zhang2023fedpetuning}. Cai et al. introduced the FedAdapter method to enhance the efficiency of adapter based FedFT schemes \cite{cai2022fedadapter}. While these methods address many challenges in FedFT, they do not effectively mitigate the problem of catastrophic forgetting \cite{delange2022continual}.
Besides communication and computation efficiency, another line of research focuses on enhancing the robustness of FedFT under non-IID data and resource heterogeneity. Heroes\cite{liu2025adaptive} proposes adaptive local updates and tensor composition to accommodate heterogeneous client capabilities. STAR\cite{liu2024enhancing} adopts progressive model expansion and confidence-aware pseudo-labeling to support training under semi-supervised and resource-constrained settings.

Existing research on catastrophic forgetting has mainly focused on centralized learning scenarios, proposing mitigation strategies \cite{delange2022continual}. For example, Hayes et al. attempted to reduce forgetting by retaining and replaying past task data \cite{hayes2020remind}, but due to FL's privacy constraints, clients cannot directly share data, making this approach impractical. Liu et al. constrained the updates of important parameters to protect knowledge from previous tasks \cite{liu2018rotate}. However, in FL environments where data is non-IID, estimating parameter importance is often unstable, limiting the effectiveness of this method \cite{karakida2019universal}. Dou et al. introduced independent parameter modules or sparse activation mechanisms for different tasks to isolate knowledge \cite{dou2023loramoe}, but these approaches significantly increase computational and storage overhead, reducing feasibility on resource constrained devices.

\section{Conclusion}\label{sec:conclusion}
This work tackles catastrophic forgetting and heterogeneity in FedFT of large language models. We propose a gradient-guided layer expansion framework that selectively augments task-critical layers while preserving pre-trained knowledge, thereby mitigating forgetting and enhancing knowledge retention. To further adapt to client-specific data and resource profiles, we design a dynamic task allocation strategy that assigns personalized training layers.
Extensive experiments demonstrate that our method not only improves the ability to resist catastrophic forgetting compared to existing methods in non-IID settings, but also enhances the convergence speed of the model.
These contributions pave the way for scalable and efficient FedFT systems. Future work will explore parameter compression and cross-modal adaptation to extend applicability to edge devices.


\bibliographystyle{IEEEtran}
\bibliography{main}

\end{document}